\documentclass[twoside]{article}

\usepackage{microtype}
\usepackage{graphicx}
\usepackage{subfigure}
\usepackage{booktabs} 
\usepackage{tikz}
\usetikzlibrary{decorations.pathreplacing}
\usetikzlibrary{fadings}
\usepackage{amsmath,amssymb}
\usepackage{wrapfig}
\usepackage{subfig}
\usepackage{wrapfig}
\usepackage{float}
\usepackage{threeparttable}
\usepackage{stfloats}
\usepackage{amssymb}
\usepackage[inline]{enumitem} 
\usepackage{bm}
\usepackage{mathtools}
\usepackage{multirow}
\usepackage{rotating}

\newcommand{\fb}[1]{\dofb#1}
\newcommand{\dofb}[1]{\textbf{#1}\nobreak\hspace{0pt}}
\usepackage{tabularx}

\usepackage[accepted]{aistats2024}
\usepackage[round]{natbib}

\usepackage{url}

\begin{document}

%
\runningtitle{NAMLSS}

%
\runningauthor{Thielmann, Kruse, Kneib, Säfken}

\twocolumn[

\aistatstitle{Neural Additive Models for Location Scale and Shape:\\
A Framework for Interpretable Neural Regression Beyond the Mean}
\aistatsauthor{Anton Thielmann\textsuperscript{*} \And René-Marcel Kruse\textsuperscript{*}}

\aistatsaddress{Clausthal University of Technology \\$^*$\footnotesize{Equal contribution} \And University of Göttingen \\$^*$\footnotesize{Equal contribution}}

\aistatsauthor{Thomas Kneib \And Benjamin Säfken}
\aistatsaddress{University of Göttingen \And Clausthal University of Technology}

]

\begin{abstract}
Deep neural networks (DNNs) have proven to be highly effective in a variety of tasks, making them the go-to method for problems requiring high-level predictive power. Despite this success, the inner workings of DNNs are often not transparent, making them difficult to interpret or understand. This lack of interpretability has led to increased research on inherently interpretable neural networks in recent years. Models such as Neural Additive Models (NAMs) achieve visual interpretability through the combination of classical statistical methods with DNNs.
However, these approaches only concentrate on mean response predictions, leaving out other properties of the response distribution of the underlying data.
We propose Neural Additive Models for Location Scale and Shape (NAMLSS), a modelling framework that combines the predictive power of classical deep learning models with the inherent advantages of distributional regression while maintaining the interpretability of additive models. The
code is available at the following link: \url{https://github.com/AnFreTh/NAMpy}

\end{abstract}

\section{Introduction}

Deep learning models have shown impressive performances on a variety of predictive tasks. 
These models represent the forefront of technology for handling unstructured data tasks, including but not limited to image classification \citep{yu2022coca, dosovitskiy2020image}, text classification \citep{huang2021balancing, lin2021bertgcn}, audio classification \citep{nagrani2021attention}, time-series forecasting \citep{zhou2022film, zeng2022transformers} and numerous other applications. 
However, the predictive performance comes not only at the price of computational demands. 
The black-box nature of deep neural networks poses hard challenges to interpretability. 
To achieve sample-level interpretability, existing methods resort to model-agnostic methods.
Locally Interpretable Model Explanations (LIME) \citep{ribeiro2016should} or Shapley values \citep{shapley1953quota} and their extensions \citep{sundararajan2020many} try to explain model predictions via local approximation and feature importance. 
Sensitivity-based approaches \citep{horel2020significance}, exploiting significance statistics, can only be applied to single-layer feed-forward neural networks and can hence not be used to model difficult non-linear effects, requiring more complex model structures.

Subsequently, high-risk domains, such as  medical applications often cannot exploit the advantages of complex neural networks due to their lack of innate interpretability. 
The creation of these innately interpretable models hence remains an important challenge. 
Achieving the interpretability from flexible statistical models as Generalized Linear Models (GLMs) \citep{nelder1972generalized} or Generalized Additive Models (GAMs) \citep{hastie2017generalized},  in deep neural networks, however, is inherently difficult.
Recently, \citet{agarwal_neural_2021} introduced Neural Additive Models (NAMs), a framework that models all features individually and thus creates visual interpretability of the single features. 
While this is an important step towards interpretable deep neural networks, any insightfulness of aspects beyond the mean is lost in the model structure.
To counter that, we propose the neural counterpart to Generalized Additive Models for Location, Scale and Shape (GAMLSS) \citep{rigby2005generalized}, the \fb{Neural} \fb{Additive} \fb{Model} for \fb{Location}, \fb{Scale} and \fb{Shape} (NAMLSS).
NAMLSS adopts and iterates on the model class of GAMLSS, in the same scope as NAMs \citep{agarwal_neural_2021} on GAMs.

The GAMLSS framework relaxes the exponential family assumption and replaces it with a general distribution family. 
The systematic part of the model is expanded to allow not only the mean (location) but all the parameters of the conditional distribution of the dependent variable to be modelled as additive nonparametric functions of the features, resulting in the following model notation:

\begin{equation*}
    \label{eq:chap1-gamlss}
    \theta^{(k)} = {g^{(k)}}^{-1} \left( \beta^{(k)} + \sum_{j=1}^{J_{k}} f_{j}^{(k)} (x_{j}^{(k)}) \right) = \eta_{\theta^{(k)}}, 
\end{equation*}

with the superscript $k = 1, \dots, K$ 
denoting the $k$-th parameter and $j=1, \dots, J$ denoting the features.

The model assumes that the underlying response observations $y_{i}$ for $i=1,2,\dots, n$ are conditionally independent given the covariates. 
The assumed conditional density can depend on up to $K$ different distributional parameters\footnote{In practice most application focus on up to four $\boldsymbol{\theta}_{i} = \left(\theta_{i}^{(1)},\theta_{i}^{(2)},\theta_{i}^{(3)},\theta_{i}^{(4)} \right)$.}. 
Each of these distribution parameters $\theta^{(k)}$ can be modelled using its additive predictor $\eta_{\theta^{(k)}}$ for $k = 1, \dots, K$, allowing for complex relationships between the response and predictor variables, as well as the flexibility to choose different distributions for different parts of the response variable.
An additional important component of the GAMLSS model is the link function $g^{(k)}(\cdot)$, which allows each parameter of the distribution vector to be conditional on different sets of covariates. 
In the case that the distribution under consideration features only one distribution parameter, the model simplifies to an ordinary GAM model.
Therefore, GAMLSS is to be seen as a conceptual extension of the GAM idea and is suitable for the extension and generalisation of approaches such as NAMs which are themselves built upon the GAM idea.
For an overview of the current state of regression models that focus on the full response distribution approaches, see \citet{kneib2021rage}.

While NAMs learn linear combinations of different input features to learn arbitrary complex functions and at the same time provide improved interpretability, these models, like their statistical counterparts GAMs, focus exclusively on modelling mean and dispersion. 
In contrast, the GAMLSS, and later the proposed NAMLSS, significantly broaden the scope by allowing all underlying parameters of the response distribution to potentially depend on the information in the covariates.

\paragraph{Contributions} The contributions of the paper hence can be summarized as follows:
\begin{itemize}
    \item We present a novel architecture for Neural Additive Models for Location, Scale and Shape.
    \item Compared to state-of-the-art GAM, GAMLSS and DNNs our NAMLSS demonstrates superior performance on benchmark datasets.
    \item We demonstrate that NAMLSS effectively captures the information underlying the data. Especially NAMLSS allows for prediction beyond point estimates, for instance prediction intervals.
    \item Lastly, we show that the NAMLSS approach allows to go beyond the mean prediction of the response and to model the entire response distribution.
\end{itemize}

\section{Literature Review} \label{chap2:literature}
The idea of generating feature-level interpretability in deep neural networks by translating GAMs into a neural framework was already introduced by \citet{potts1999generalized} and expanded by \citet{de2007generalized}. 
While the framework was remarkably parameter-sparse, it did not use backpropagation and hence did not achieve as good predictive results as GAMs, while remaining less interpretable.
More recently, \citet{agarwal_neural_2021} introduced NAMs, a more flexible approach than the Generalized Additive Neural Networks (GANNs) introduced by \citet{de2007generalized} that leverages the recent advances in the field of Deep Learning. 

NAMs are a class of flexible and powerful machine learning models that combine the strengths of neural networks and GAMs. 
These models can be used to model complex, non-linear relationships between response and predictor variables, and can be applied to a wide range of tasks including regression, classification, and time series forecasting.
The basic structure of a NAM consists of a sum of multiple components, each representing a different aspect of the relationship between the response and predictor variables. 
These components can be linear, non-linear, or a combination of both, and can be learned using a variety of optimization algorithms.
One of the key advantages of NAMs is their inherent ability to learn the interactions between different predictor variables and the response without the need for manual feature engineering. 
This allows NAMs to capture complex relationships in the data that may not be easily apparent to the human eye.

The general form of a NAM can be written as:

\begin{equation}
    \label{eq:NAM2}
    \mathbb{E}(y) = h \left( \beta + \sum_{j=1}^{J}f_j(x_j) \right),
\end{equation}

where $h(\cdot)$ is the activation function used in the output layer, $x \in \mathbb{R}^{j}$ are the input features, $\beta$ is the global intercept term, and $f_{j} : \mathbb{R} \rightarrow \mathbb{R}$ represents the Multi-Layer Perceptron (MLP) corresponding to the $j$-th feature.
The similarity to GAMs is apparent, as the two frameworks mostly distinguish in the form the individual features are modelled. 
$h(\cdot)$ is comparable to the link function $g(\cdot)$.

Several extensions to the NAM framework have already been introduced. It is possible to take into account pairwise or higher order interaction effects \citep{yang2021gami, enouen2022sparse, wang2021partially, dubey2022scalable}. 
\citet{chang2021node} introduced NODE-GAM, a differentiable model based on oblivious neural decision trees developed for high-risk domains.
All these models follow the additive framework from GAMs and learn the nonlinear additive features with separate networks, one for each feature or feature interaction, either leveraging MLPs \citep{potts1999generalized, de2007generalized, agarwal_neural_2021, yang2021gami, radenovic2022neural}, using decision trees \citep{chang2021node} or using Splines \citep{rugamer2020semi, Seifert2022penalized, luber2023structural}.

The applications of such models range from nowcasting \citep{jo2022neural}, financial applications \citep{chen2022generalized}, to survival analysis \citep{peroni2022extending}. 
While the linear combination of neural subnetworks provides a visual interpretation of the results, any interpretability beyond the feature-level representation of the model predictions is lost in their black-box subnetworks.

\section{Beyond the Mean}

Obviously, the mean (or the arithmetic mean as its empirical counterpart) provides only a rather incomplete description of a probability distribution (the empirical distribution of corresponding observations, in case of the arithmetic mean). While this fact is widely acknowledged when it comes to exploratory data analysis, it is also widely ignored in the context of prediction models where the focus is typically on predicting expected outcomes. This narrow focus reflects an interest in common or average observations, but is misleading when phenomena such as risk, extremes, or uncertainty are central to an analysis. With the GAMLSS-based framework considered in this paper, we are able to quantify effects of covariates not only on the mean, but on any parameter of a potentially complex distribution assumed for the responses. As major advantage, the resulting models can determine changes in all aspects of the response distribution, such as variance, skewness or tail probabilities. This also contributes to properly disentangling aleatoric from epistemic uncertainty.

\begin{figure}
    \centering
    \includegraphics[width=0.5\textwidth]{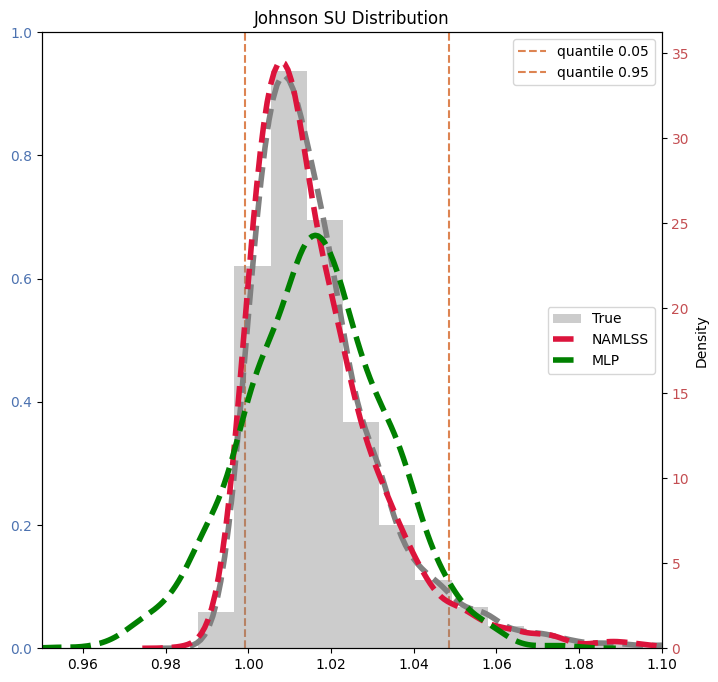}
    \caption{\textbf{Johnson's S$_U$ Distribution:} Simulated Johnson's S$_U$ distribution and the fit of a simple NAMLSS (see Figure \ref{fig:namlss_structure1}) and a MLP. While the MLP achieves an impressive fit concerning the quadratic loss, it clearly cannot capture the underlying distribution adequately.}
    \label{Fig:JohnsonSU}
\end{figure}

Changing the focus from regression models for the mean to regression for distributions also requires changes in the evaluation metric that is used to compare rivalling model specifications. 
More precisely, the evaluation metric should be proper \citep{GneRaf2007}, i.e. enforce the analyst to report their true beliefs in terms of a predictive distribution. While the MSE that is commonly employed in mean-based modelling is proper for the mean, it is not for general distributions. We therefore will rely on the negative log-likelihood (also refereed to as the log-score) as a proper score for comparing distributional regression models (see Supplemental Material for details).
Additionally to the negative log-likelihood, we use the Continuous Ranked Probability Score \citep{GneRaf2007} for model evaluation, given by:
$$CRPS(F, x) = -\int_{-\infty}^{\infty} (F(y) - \mathbf{1}_{y \geq x})^2 \, dy.$$ 
Note that the CRPS is also defined for models not specifically predicting all distributional parameters and thus allows a fair distributional comparison for all tested models: 
$$CRPS(p, x) = \frac{1}{2} \left( \mathbb{E}_{p|X} \left[ |X - \mathbf{X} \right| - \mathbb{E}_{p|X} \left[ |X - x| \right] \right).$$ See \cite{GneRaf2007} for more details.

While predicting all parameters from a distribution may not always improve predictive power, understanding the underlying data distribution is crucial in high-risk domains and can provide valuable insights about feature effects. As an example, Figure \ref{Fig:JohnsonSU} illustrates the fit of our approach on data following a Johnson's S$_U$ distribution, including 3 features, compared to the fit of a MLP that minimizes the Mean Squared Error (MSE). 
The MLP has a better predictive performance with an MSE of 0.0002, however, NAMLSS is able to reflect the underlying data distribution much more accurately (as shown in Figure \ref{Fig:JohnsonSU}), even though it has an MSE of 0.0005.

The idea of focusing on more than the underlying mean prediction is thus certainly relevant and has been an important part, especially of the statistical literature in recent years. There has been a strong focus on the GAMLSS \citep{rigby2005generalized} framework, conditional transformation models \citep{hothorn2014conditional}, density regression \citep{wang1996mixed} or quantile and expectile regression frameworks. 
However, these methods are inferior to machine and deep learning techniques in terms of pure predictive power; the disadvantage of not being able to deal with unstructured data forms such as images, text or audio files; or the inherent problems of statistical models in dealing with extremely large and complex data sets.
One resulting development to deal with these drawbacks is frameworks that utilize statistical modelling methods and combine them with machine learning techniques such as boosting to create new types of distributional regression models such as boosted generalized additive model for location, scale and shape as presented by \citet{hofner2014gamboostlss}. 
More recently, there has been an increasing trend in incorporating distributional strategies within black-box boosting frameworks, often by modifying the model's objective function  (e.g. \citep{duan2020ngboost, marz2022distributional}), through ensemble methods \citep{malinin2020uncertainty} or by leveraging normalizing flows \citep{wielopolski2022treeflow}.
However, the models leveraging boosting techniques, while successfully modelling all distributional parameters, lack the inherent interpretability from GAMLSS or even the visual interpretability from NAMs. Consequently, the benefits of analyzing individual feature contributions to distributional parameters, such as the variance in a normal distribution, may not be fully realized.

\section{Methodology}
\label{chap3:methodology}

While NAMs incorporate some feature-level interpretability and hence entail easy interpretability of the estimated regression effects, they are unable to capture skewness, heteroskedasticity or kurtosis in the underlying data distribution due to their focus on mean prediction.
Therefore, the presented method is the neural counterpart to GAMLSS, offering the flexibility and predictive performance of neural networks while maintaining feature-level interpretability and which allows estimation of the underlying total response distribution.

Let $\mathcal{D} = \{ (x^{(i)}, y^{(i)})\}_{i=1}^n$ be the training dataset of size n. Each input $x = (x_1, x_2, \dots, x_J)$ contains J features. 
$y$ denotes the target variable and can be arbitrarily distributed. NAMLSS are trained by minimizing the negative log-likelihood as the loss function, $ -\log \left(\mathcal{L}(\theta|y)\right)$ by optimally approximating the distributional parameters, $\theta^{(k)}$. Each parameter, $\theta^{(k)}$, is defined as: 

\begin{equation}\label{eq:NAMLSS}
    \theta^{(k)} = h^{(k)} \left( \beta^{(k)} + \sum_{j=1}^{J} f_j^{(k)}(x_j)\right),
\end{equation} 

\begin{figure*}
\centering
\includegraphics[width=0.65\textwidth]{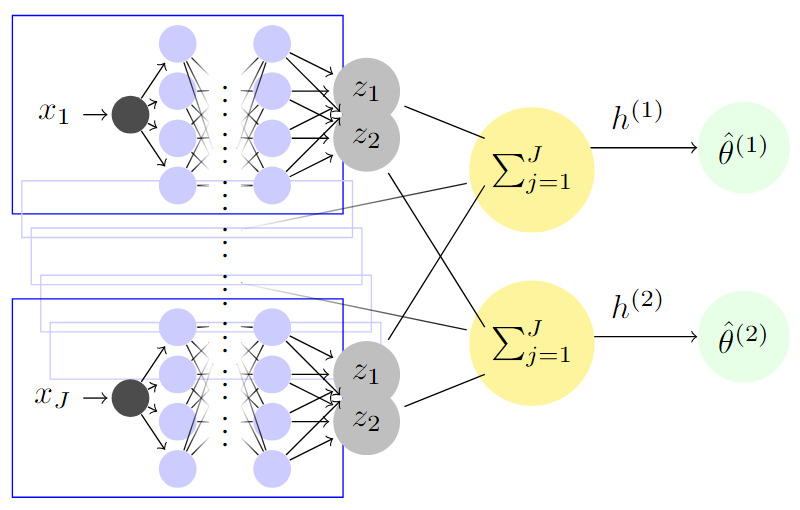}
\caption{\small{The network structure of a NAMLSS model. 
Each input variable is handled by a different neural network with $k$ outputs for each distributional parameter. 
$h^{(k)}$ are different activation functions depending on the distributional parameter that is modelled. 
E.g. a quadratic transformation for modelling the variance in a normally distributed variable to ensure the non-negativity constraint. 
The presented structure demonstrates a NAMLSS modelling a distribution with two parameters, e.g. a normal distribution.}}

\label{fig:namlss_structure2}
\end{figure*}

where $h^{(k)}(\cdot)$ denotes the output layer activation functions dependent on the underlying distributional parameter, $\beta^{(k)}$ denotes the parameter-specific intercept and $f_{j}^{(k)} : \mathbb{R} \rightarrow \mathbb{R}$ represents the feature network for parameter $k$ for the $j$-th feature, subsequently called the \textit{parameter-feature network}.

Just as in GAMLSS, $\theta^{(k)}$ can be derived from a subset of the $J$ features, however, due to the inherent flexibility of the neural networks, defining each $\theta^{(k)}$ over all $J$ is sufficient, as the individual feature importance for each parameter, $\theta^{(k)}$, is learned automatically.
Each parameter-feature network, $f_j^{(k)}$, can be regularized employing regular dropout coefficients in conjunction with feature dropout coefficients, $\lambda_{1j}^{(k)}$ and $\lambda_{2j}^{(k)}$ respectively, as also implemented by \citet{agarwal_neural_2021}.
%

We propose two different network architectures that can both flexibly model all distributional parameters. 
The first model architecture, possible due to the flexibility of neural networks is depicted in Figure \ref{fig:namlss_structure2} and creates $J$ subnetworks, with each subnetwork having a $K$-dimensional output layer. This architecture thus creates the same number of subnetworks as a common GAM and differs from the classical GAMLSS architecture as it comprises less feature networks/shape functions.
Each distributional parameter, $\theta^{(k)}$, is subsequently obtained by summing over the $k$-th output of the $J$ subnetworks. 
Every dimension in the output layer can be activated using different activation functions, according to parameter restrictions. 
This allows the capture of interaction effects between the distributional parameters in each of the subnetworks.
Equation \ref{eq:NAMLSS} would only slightly be adjusted, to account for the subnetwork $f_j$ now mapping to $\mathbb{R}^k$,  
$f_{j} : \mathbb{R} \rightarrow \mathbb{R}^k$:
\begin{equation}
    \theta^{(k)} = h^{(k)} \left( \beta^{(k)} + \sum_{j=1}^{J} f_j^{}(x_j)\left[:, k\right]\right),
\end{equation} \label{eq:NAMLSS2}
with $\left[:, k\right]$ denoting an index and representing the $k-th$ index of $f_{j} : \mathbb{R} \rightarrow \mathbb{R}^k$.
Note, that the superscript $^{(k)}$ is missing from the subnetwork 
$f_j$, as only $J$ subnetworks are trained.
For instance, consider a simple example of a normal distribution: $\hat\mu = h^{(0)} + \beta^{(0)} \sum_{j=1}^{J} f_j(x_j)[:, 0]$ (with Pythonic indexing where 0 is the first index), and $\hat\sigma = g\left(h^{(1)} + \beta^{(1)} \sum_{j=1}^{J} f_j(x_j)[:, 1]\right)$, where $g(\cdot)$ describes the softplus activation function, $g(x) = \log(1 + e^x)$ since $\sigma \in \mathbb{R^+}$. For $\mu$, $g(\cdot)$ is simply the linear activation and thus neglectable as $\mu \in \mathbb{R}$. These predictions are directly utilized in the loss function and thus serve as parameters for the negative log-likelihood, which, for the normal distribution, is given by: 
$$-1 \cdot \log \left(\mathcal{L}(\hat\mu, \hat\sigma^2 | y)\right) = \frac{n}{2} \log(2\pi\hat\sigma^2) + \frac{1}{2\hat\sigma^2} \sum_{i=1}^n (y_i - \hat\mu)^2.$$ 
The negative log-likelihood is then minimized using gradient descent, similar to standard neural networks. The additivity constraint serves two important purposes: First, as an additional regulator against overfitting, and second, to ensure interpretability. This constraint allows to visualize all feature effects on all distributional parameters. In terms of interpretability, accounting for uncertainty can be highly beneficial, particularly in domains such as healthcare and biology, where the impact of feature variance on outcomes is critical.

The second proposed architecture is depicted in Figure \ref{fig:namlss_structure1} in the Supplemental Material and creates $J$ subnetworks for each of the $K$ distributional parameters and thus much more resembles the classical GAMLSS architecture\footnote{Note, that for distributions where only one parameter is modelled, the two proposed NAMLSS structures are identical.}.
Each distributional subnetwork is comprised of the sum of the parameter-feature networks $f_{j}^{(k)}$.  
Hence we create $K\times J$ parameter-feature networks as denoted in equation \ref{eq:NAMLSS}. 
To account for distributional restrictions, each distributional subnetwork is again specified with possibly differing activation functions in the output layer.

\begin{figure*} 
\centering
\includegraphics[width=0.85\textwidth]{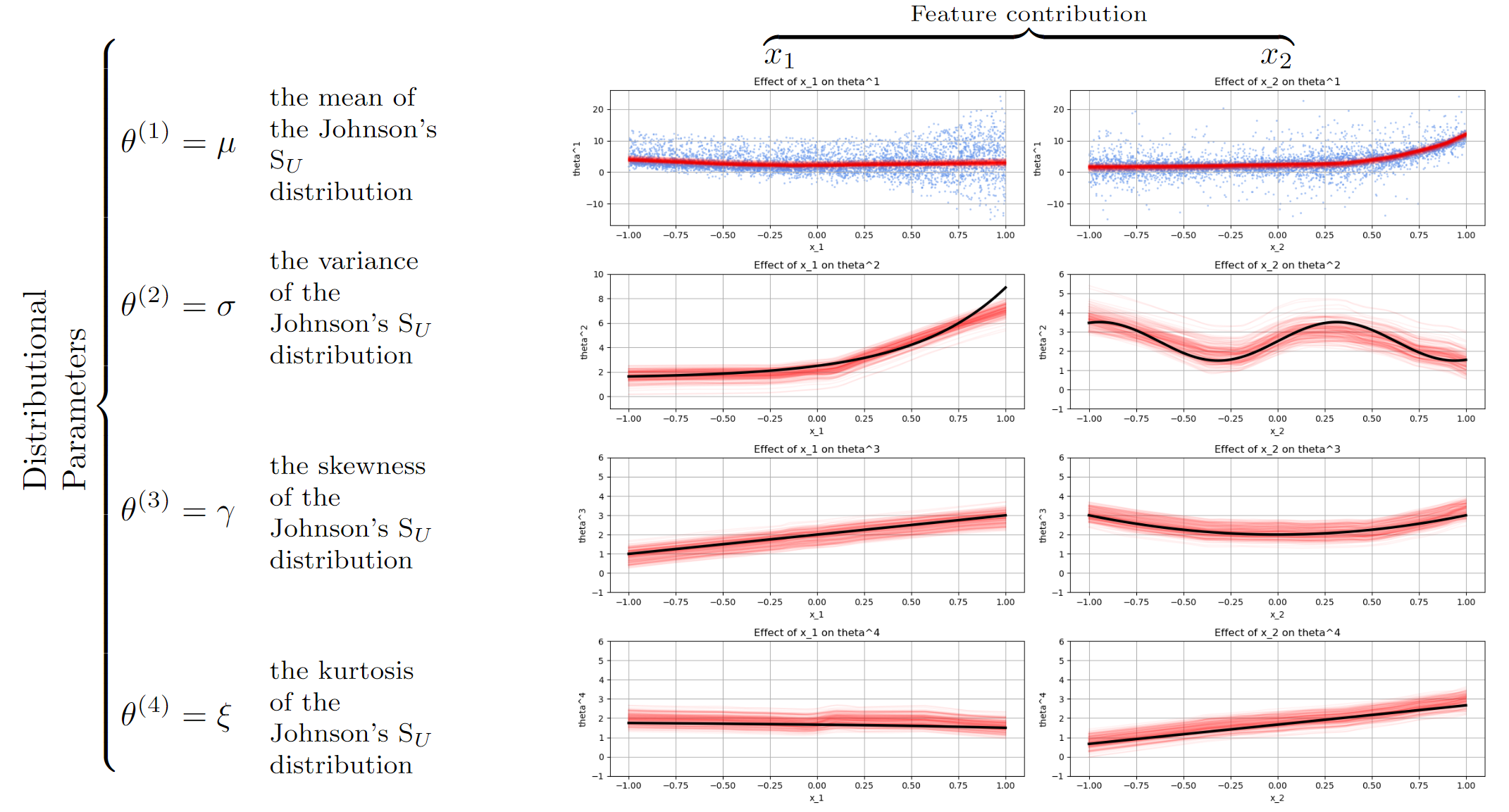} 
    \caption{\small{NAMLSS distributional parameter prediction for a Johnson's S$_U$ Distribution over 150 runs. NAMLSS accurately detects the feature effects of $x_1$ and $x_2$ on all distributional parameters, $\theta^{(1)}$ the mean, $\theta^{(2)}$ the standard deviation, $\theta^{(3)}$ the skewness and $\theta^{(4)}$ the tailweight.}}
    \label{ablation}
\end{figure*}

Figure \ref{ablation} demonstrates the accurate parameter fit of NAMLSS for a Johnson's S$_U$ distribution. Each parameter is depicted in a different row. The black line shows the data generating functions. The simulated feature effects (columns) are accurately captured over all of the 150 runs for all 4 distributional parameters.

\section{Benchmarking}\label{chap4:benchmarking}
To demonstrate the competitiveness of the presented method, we perform several analyses.
First, we compare NAMLSS with the most common statistical distributional regression approach GAMLSS \citep{stasinopoulos2000modelling}.
\paragraph{Synthetic data comparison study} 

The synthetic data used for this task is generated from the same underlying processes. Five features are included in each application. The data-generating functions used to generate the true underlying distributional parameters can be found in the Supplemental Material \ref{app:daten_simulation_formel}.
Each of the five input vectors $x_j$ is sampled from a uniform distribution $\mathcal{U}(0,1)$, with a total of $n=3000$ observations per data set. 
The remaining parameters are generated based on the input vectors and the chosen distribution. 
We selected distributions that are widely used, popular in science, or relatively complex to reflect a diverse range of scenarios. 
The results can be found in Table \ref{tab:simulation_study}.

We find that the presented NAMLSS outperforms GAMLSS for all distributions except the Poisson distribution. 
This can be attributed to the fact that the Poisson distribution only involves a single distributional parameter.

%
%
%

\begin{table}
    \centering
    \small
    \caption{\small{\textbf{Results for Synthetic data:} We compare NAMLSS with the baseline of additive distributional models, GAMLSS.}}

        \begin{tabular}{l c c }
        \toprule
        Distribution & GAMLSS & NAMLSS \\
        \midrule
        & \multicolumn{2}{c}{Neg Log-Likelihood $\downarrow$}                   \\
        \cmidrule(r){2-3}
        Binomial        & \textbf{397} & 274   \\
        Poisson         & 800 & \textbf{802}  \\
        Normal          & \textbf{600} & 589   \\
        Inv. Gaussian   & \textbf{385} & 377   \\
        Weibull         & \textbf{625} & 621   \\
        Johnson's S$_U$ & \textbf{370} & 326   \\
        Gamma           & \textbf{426} & 410   \\
        Logistic        & \textbf{731} & 682   \\

        \bottomrule
        \end{tabular}

    \label{tab:simulation_study}
\end{table}

\paragraph{Experiments with Real World Data}

We compare the performance of NAMLSS  with several state-of-the-art models including neural as well as non-neural approaches and orientate on the benchmarks performed by \citet{agarwal_neural_2021}. 

Additionally, we compare related methods of distribution-focused data analysis approaches that overcome the focus on relating the conditional mean of the response to features and instead target the complete conditional response distribution. Note, that we include the mean-centric models as a method to demonstrate the advantages of distributional approaches, particularly in terms of proper scoring metrics (negative log-likelihood, CRPS). Where no closed-form solution for the CRPS exists (Inverse Gamma distribution), we report the Kullback-Leibler Divergence.
We have selected the following baselines for the comparisons:

The Multilayer Perceptron (MLP).
Gradient Boosted Trees (XGBoost), based on decision tree-based gradient boosting using the implementation provided by \citep{Chen:2016:XST:2939672.2939785}.
Neural Additive Models (NAMs), represented as a linear combination of DNNs as described in equation (\ref{eq:NAM2}) and presented by \citep{agarwal_neural_2021}.
Explainable Boosting Machines (EBMs), which are state-of-the-art Generalized Additive Models leveraging shallow boosted trees \citep{nori2019interpretml}.
Neural Generalized Additive Model (NodeGAM), leveraging Neural Oblivious Decision Trees \citep{chang2021node}.
The Deep Distributional Neural Network (DDNN), a fully connected neural network trained to minimize the negative log-likelihood of the specified distribution. 
GAMLSS, employing standard GAMLSS models using the R implementation from \citep{rigby2005generalized}.
gamboost for Location Scale and Shape (gamboostLSS), fitting GAMLSS by employing boosting techniques as proposed by \citep{hofner2014gamboostlss}.

These baseline models provide a diverse set of techniques for our comparative analysis.
\begin{table}
\centering
\begin{threeparttable}
\caption{\small{Average Rank table: The average rank over all models over all real world datasets. Both NAMLSS architectures outperform the mean reducing models by a large margin. We find the second NAMLSS architecture that captures parameter interaction effects to outperform the vanilla additive structure.}}

    \begin{tabularx}{0.5\textwidth}{X c c}
    \toprule
               
     Model & Avg. Rank LL & Avg. Rank CRPS/KL  \\ 
        \midrule
        MLP             & 7.4 & 6.8 \\
        XGBoost         & 8.2 & 7.4 \\
        NAM             & 7.4 & 7.5 \\
        EBM             & 6.5 & 6.1 \\
        NodeGAM         & 7.8 & 7.5 \\
        DDNN            & 4.0 & 2.1 \\
        GAMLSS          & 4.5 & 6.0 \\
        gamboostLSS     & 4.2 & 5.3 \\
        \midrule
        NAMLSS\tnote{1} & 1.9 & 3.1 \\
        NAMLSS\tnote{2} & \textbf{1.6} &  \textbf{1.9}  \\
        \bottomrule
    \end{tabularx}
    \label{tab:rankings}
     \begin{tablenotes}
      \small
      \begin{enumerate*}     
      \item[$^{1}$] With $J \times K $ subnetworks\\ 
      \item[$^{2}$] With $J$ subnetworks\\ 
      \end{enumerate*}
    \end{tablenotes}
    \end{threeparttable}  
\end{table}

We preprocess all used datasets exactly as done by \citet{agarwal_neural_2021}.
We perform 5-fold cross-validation for all datasets and report the average performances over all folds as well as the standard deviations. 
For reproducability, we have only chosen publicly available datasets. 
The datasets, as well as the preprocessing and the seeds set for obtaining the folds, are described in detail in the Supplemental Material, \ref{app:reproducability}.
We fit all models without an intercept and explicitly do not model feature interaction effects. 

For datasets following a Gaussian distribution  we use the California Housing (CA Housing) dataset \citep{pace1997sparse} from sklearn \citep{pedregosa_scikit-learn_2011}, the Insurance dataset \cite{lantz2019machine}, the Abalone dataset \citep{Dua:2019} and standard normalize the response variables. Thus, a normal distribution $\mathcal{N}\left(\mu, \sigma^{2} \mathcal{I} \right)$ of the underlying response variables is assumed. This notably serves the objective of illustrating that, even in scenarios where mean-centric models designed to minimize the MSE should theoretically excel, they fall short in comparison to distributional approaches\footnote{As the (negative) log-likelihood of a normal distribution (see equation (\ref{app:normal})) is dependent on two parameters, but models as an MLP or XGBoost only predict a single parameter, we adjust the computation accordingly and use the true standard deviation calculated from the underlying data for XGBoost, EBM, NAM and MLP.}.
For a (binary) classification benchmark we use the FICO dataset \citep{FICO}, the Shrutime dataset and the Telco dataset. A logistic distribution, $\mathcal{LO}\left(\mu, s \right)$, of the underlying response variable was assumed (see equation (\ref{app:logistic}) for the log-likelihood).  Again, we use the true standard deviation of the underlying data for the models only resulting in a mean prediction. 
For the Melbourne and Munich datasets, also analyzed by \citet{rugamer2020semi}, we assume an Inverse Gamma distribution $\mathcal{IG}(\alpha, \beta)$ as the underlying data distribution (see equation (\ref{app:inverse-gamma}) for the log-likelihood)\footnote{See Supplemental Material for further details on activation functions.}. 

\begin{table*}
\centering
\small
       \caption{\small{\textbf{Benchmark results:}
       For models not explicitly modelling a shape parameter, the shape is approximated with a constant as the true standard deviation of the dependent variable. 
       Lower negative log-likelihoods ($\ell$) are better. 
       We report results on 6 commonly used datasets (see Supplemental Material for further results).
       The California Housing dataset for predicting house prices \citep{pace1997sparse}, an Insurance dataset for predicting billed medical expenses \citep{lantz2019machine}, the Abalone dataset for predicting number of rings in trees \citep{Dua:2019}, two AirBnb datasets and the FICO dataset for predicting \textit{Risk Performance}.}}
\begin{threeparttable}
    \begin{tabular}{l c c  c c c c}
    \toprule
    & \multicolumn{6}{c}{Negative Log-Likelihood $\ell$ ($\downarrow$)}  \\
     & \multicolumn{3}{c}{\textbf{Normal}} &  \multicolumn{2}{c}{\textbf{Inv. Gamma}} & \textbf{Logistic} \\
               
     Model & CA Housing & Insurance & Abalone &  Munich & Melbourne & FICO  \\ 
        
        & & & & \\
        \midrule
        MLP             & 4191 \footnotesize{$\pm$(42)}               & 266.8 \footnotesize{$\pm$ (11)}         &    966.2  \footnotesize{$\pm$(27)}          & 6827 \footnotesize{$\pm$ (178)}            &  22999  \footnotesize{$\pm$ (232)}    & 1813 \footnotesize{$\pm$(6)}             \\
        XGBoost         & 4219 \footnotesize{$\pm$(40)}               & 266.8 \footnotesize{$\pm$  (9)}         &    982.0  \footnotesize{$\pm$(33)}          & 5618 \footnotesize{$\pm$ (152)}            &  20471 \footnotesize{$\pm$ (242)}     & 1976 \footnotesize{$\pm$(13)}             \\
        NAM             & 4251 \footnotesize{$\pm$(43)}             & 474.7 \footnotesize{$\pm$ (73)}           &  956.8   \footnotesize{$\pm$(22)}           &  5892 \footnotesize{$\pm$ (37)}            &  25375  \footnotesize{$\pm$ (844)}    & 1809 \footnotesize{$\pm$ (8)}              \\
        EBM             & 4202 \footnotesize{$\pm$(42)}               & 263.8 \footnotesize{$\pm$ (10)}         &   965.1  \footnotesize{$\pm$(22)}           &  5474 \footnotesize{$\pm$ (56)}            &  20361  \footnotesize{$\pm$ (207)}    & 1944 \footnotesize{$\pm$(21)}          \\
        NodeGAM         & 4206 \footnotesize{$\pm$(89)}               & 279.1 \footnotesize{$\pm$ (11)}         &   958.3  \footnotesize{$\pm$(23)}           &  5984 \footnotesize{$\pm$ (135)}           &  21896  \footnotesize{$\pm$ (261)}    & 1942 \footnotesize{$\pm$(21)}             \\
        \hline
        DDNN             & 2681 \footnotesize{$\pm$(1279)}              & 178.2 \footnotesize{$\pm$ (30)}        &    897.2 \footnotesize{$\pm$(159)}          & 5555 \footnotesize{$\pm$ (34)}            &   20790   \footnotesize{$\pm$ (29)}   & 1230 \footnotesize{$\pm$ (48)}        \\
        GAMLSS           & 3512 \footnotesize{$\pm$(67)}               & 175.5 \footnotesize{$\pm$ (28)}         &    870.8 \footnotesize{$\pm$(16)}            & 5419 \footnotesize{$\pm$ (61)}           &  26353    \footnotesize{$\pm$ (45)}   & 1321 \footnotesize{$\pm$ (30)}           \\
        gamboostLSS      & 3812   \footnotesize{$\pm$(52)}             & {173.0} \footnotesize{$\pm$ (28)}       & 815.1 \footnotesize{$\pm$ (29)}              & 5421 \footnotesize{$\pm$ (33)}           &   26436    \footnotesize{$\pm$ (48)}  & 1191 \footnotesize{$\pm$ (30)}             \\
        \midrule
        NAMLSS\tnote{1} &  2667 \footnotesize{$\pm$ (91)}             & 172.7 \footnotesize{$\pm$ (23)}          &      869.8 \footnotesize{$\pm$(118)}         & \textbf{5383} \footnotesize{$\pm$ (24)}  &   \textbf{19517}   \footnotesize{$\pm$ (68)}  &  1201 \footnotesize{$\pm$ (41)}    \\  
        NAMLSS\tnote{2} &  \textbf{2329} \footnotesize{$\pm$ (176)}  &  \textbf{172.6} \footnotesize{$\pm$ (20)} & \textbf{802.3} \footnotesize{$\pm$(41)}         &  5422 \footnotesize{$\pm$ (22)}       &   19675      \footnotesize{$\pm$ (67)}         &  \textbf{1160} \footnotesize{$\pm$ (49)}    \\
        \bottomrule
    \end{tabular}
     \begin{tablenotes}
      \small
      \begin{enumerate*}
      
      \item[$^{1}$] With $J \times K $ subnetworks. See Table \ref{fig:namlss_structure1} for an exemplary network structure.\\ 
      \item[$^{2}$] With $J$ subnetworks and each subnetwork returning a parameter for the location and shape respectively. See Table \ref{fig:namlss_structure2} for an exemplary network structure.\\ 
      \end{enumerate*}
    \end{tablenotes}
    \end{threeparttable}

    \label{benchmark:1}
\end{table*}

The NAMLSS approach achieves the lowest negative log-likelihood and CRPS values for all of the datasets as shown in table \ref{tab:rankings}. The architecture that implicitly captures the distributional parameter interactions slightly outperforms the architecture more closely related to GAMLSS.

\begin{figure}
    \centering
    \begin{subfigure}
        \centering
        \includegraphics[width=0.48\linewidth]{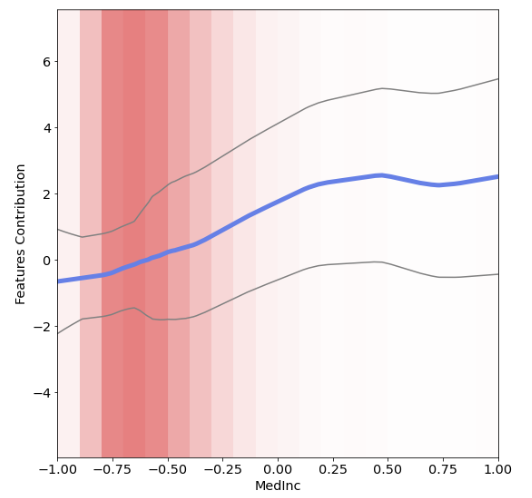}%
        \hfill
        \includegraphics[width=0.48\linewidth]{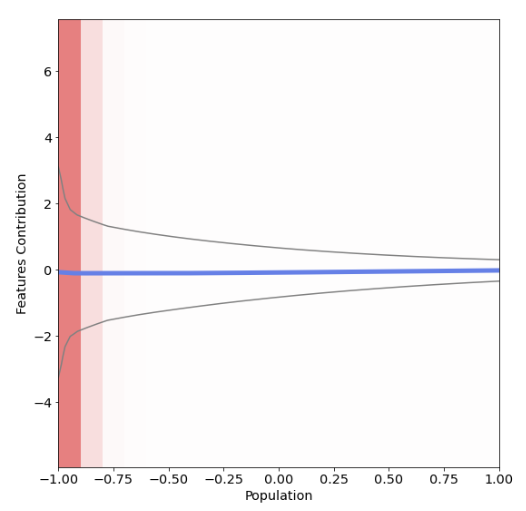}
    \end{subfigure}
    \caption{\small{\textbf{California Housing:} Graphs for median income and 
    population respectively learned by the NAMLSS model. We see an increase in housing prices with a larger median income. Additionally, we find a larger variance in housing prices in less densely populated areas.}
    \label{Fig:medinc}}
\end{figure}

\begin{figure}[H]
  \centering
  \includegraphics[width=0.48\textwidth]{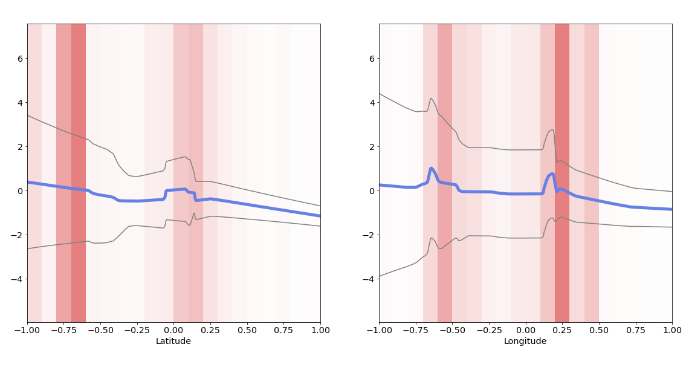}
  \caption{\small{\textbf{California Housing:} Graphs for longitude and latitude respectively learned by the NAMLSS model. NAMLSS captures changes in mean as well as variance. Therefore, the plotted standard deviations change in dependence of the longitude and latitude. The house price jumps around the location of Los Angeles are depictable. Additionally, we find a decrease in variance for areas further away from the large cities.}}
  \label{Fig:long_lat}
\end{figure}

One of the strengths of the NAMLSS, in contrast to the DNNs, is its interpretability at the feature level. 
Similar to NAMs, we can plot and visually analyze the results (see Figures \ref{Fig:medinc} and \ref{Fig:long_lat}). 
Additionally, we are able to accurately depict shifts in variance in the underlying data. 
It is, for example, clearly distinguishable, that with a larger median income, the house prices tend to vary much stronger than with a smaller median income (see Figure \ref{Fig:medinc}). 
A piece of information, that is lost in the models focusing solely on mean predictions
Additionally, we are capable of accurately representing sharp price jumps around the location of San Francisco, depicted by the jumps in the graphs for longitude and latitude (see Figure \ref{Fig:long_lat}) as compared to GAMLSS, NAMLSS are additionally capable of representing jagged shape functions.

\section{Conclusion \& Future Work}\label{chap5:interpretability}
We have presented Neural Additive Models for Location, Scale and Shape and their theoretical foundation as the neural counterpart to GAMLSS. 
NAMLSS can model an arbitrary number of parameters of the underlying data distribution while preserving the predictive quality of NAMs. 
The visual intelligibility achieved by NAMs is also maintained by NAMLSS, with the added benefit of gaining further insights from knowledge of additional distribution characteristics.
Hence, NAMLSS are a further step in the direction of fully interpretable neural networks and already offer interpretability that may make them suitable for high-risk domains.

The extensibility of NAMLSS offers many different further applied and theoretical research directions. 
One important point is the extension of the modelling of the distribution of the response variable.  
Many empirical works focus on modelling not just one, but several responses conditionally on covariates.  
One way to do this is to use copula methods, which are a valuable extension of our approach, hence including a copula-based approach for NAMLSS models would greatly improve the overall general usefulness. 
Another possible extension would be the adaptation to mixture density networks, as e.g. done by \citet{Seifert2022penalized}.
Another possible focus is to switch our approach to a Bayesian-based training approach. 
Bayesian approaches are particularly well suited to deal with epistemic uncertainty and to incorporate it into the modelling. 
Another advantage is that Bayesian approaches are particularly suitable in cases where insufficiently small training datasets have to be dealt with and have been shown to have better prediction performance in these cases.
 
NAMLSS, akin to NAM, EBM and Node-GAM initially centers on tabular data. However, the model seamlessly extends its capability to encompass multimodal data by incorporating components such as a CNN for images as one of the feature networks. In this context, each distributional parameter is modelled by: $\theta^{(k)} = g\left(h^{(k)} + \beta^{(k)} \sum_{j=1}^{J} f_j(x_j)[:, k] + f_{img}(Z)[:, k]\right)$, where $Z$ represents an image input and $f_{img}(Z)$ denotes e.g. a CNN. Further identifiability constraints such a orthogonalization \citep{rugamer2023ortho} could be added to account for identifiability of the image effects.

\section{Limitations}

Although the presented method of NAMLSS takes advantage of the interpretation capabilities of the NAM framework and thus offers a better and easier interpretation of the results compared to pure deep learning approaches, it is still beholden to classical statistical models with their inherent interpretability and explainability.
A critical point in the application of our proposed method, as well as comparable distributional statistical methods, is the choice of the correct distributional assumptions. The choice of the assumed distribution can strongly influence the results of the model. Our approach requires some basic mathematical-statistical knowledge from the user. Also, the understanding that the presented approach focuses on (log)-likelihood and thus deviates from the classical approach of simply minimising an error measure may require some users to rethink their understanding of the model results.

\section*{Acknowledgements}
Funding by the Deutsche Forschungsgemeinschaft (DFG, German Research Foundation) within project 450330162 is gratefully acknowledged.


\newpage
\bibliography{bib.bib}
\bibliographystyle{apalike}

\onecolumn
\section{Supplemental Material for NAMLSS}
\label{chapA:appendix}

\subsection{Network architecture}
\begin{figure}[H]
\centering
\includegraphics[width=0.5\textwidth]{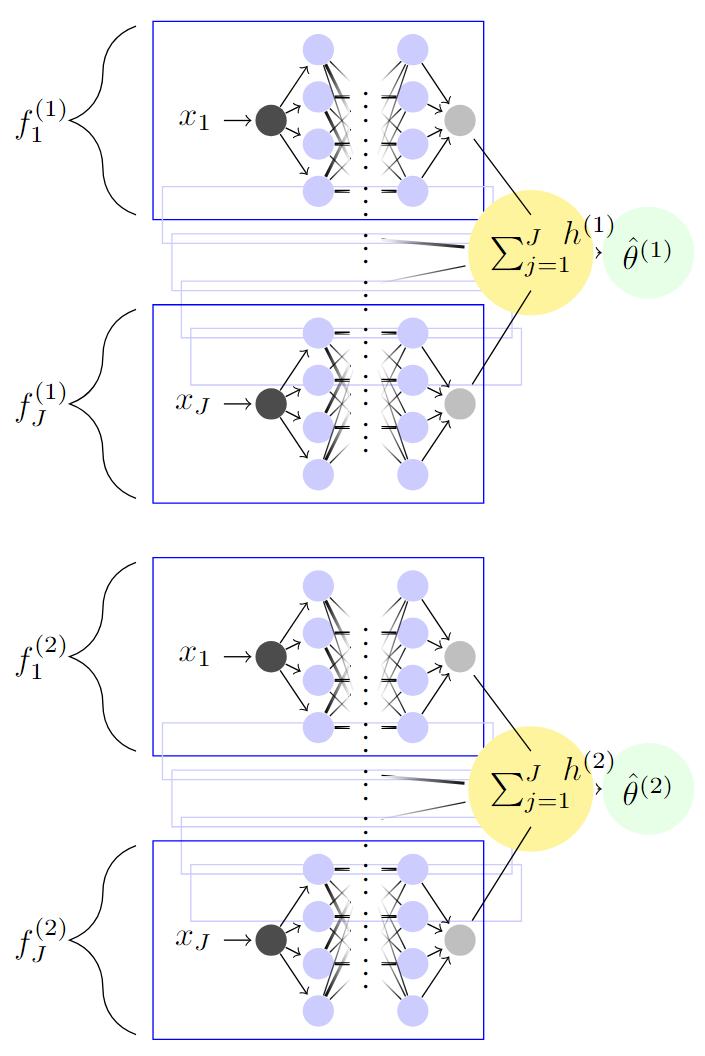}
\caption{\footnotesize{The network structure of a simple NAMLSS model. Each input variable as well as each distributional parameter is handled by a different neural network. $h_k$ are different activation functions depending on the distributional parameter that is modelled. E.g. a quadratic transformation for modelling the variance in a normally distributed variable to ensure the non-negativity constraint.}}
\label{fig:namlss_structure1}
\end{figure}

We propose two different network architectures that can both flexibly model all distributional parameters. One is depicted in Figure \ref{fig:namlss_structure1} and creates J subnetworks for each distributional parameter. Each distributional subnetwork is comprised of the sum of  $f_{j}^{(k)}$.  Hence we create $K\times J$ subnetworks. To account for distributional restrictions, each distributional subnetwork is specified with possibly differing activation functions in the output layer.

The second model architecture is depicted in Figure \ref{fig:namlss_structure2}. Here we only create $J$ subnetworks and hence have the same amount of subnetworks as a common NAM. Each subnetwork then has a $k$-dimensional output layer. Each distributional Parameter, $\theta^{(k)}$, is subsequently obtained by summing over the $k$-th output of the $J$ subnetworks. Each dimension in the output layer can be activated using different activation functions, adjusting to parameter restrictions.

\newpage
\subsection{Scoring}
We use negative-log likelihoods as well as the CRPS for evaluation metrics. All used log-likelihoods are given in the following.

\subsubsection{Log-Likelihoods}\label{Appendix:ll}
As the presented method minimizes negative log-likelihoods, we created a comprehensive list of all the log-likelihoods of the distributions used in the paper.
When we reference the results of NAMLSS these are the log-likelihoods we used for fitting the models as well as evaluating them.

\paragraph{(Bernoulli) Logistic Distribution}
The log-likelihood function for a logistic distribution is given by:
\begin{equation*}
    \log\left( \mathcal{L}(\mu, \sigma | y)\right) 
      = \sum_{i=1}^n \left[ y_i\log(\frac{1}{1+e^{-(\frac{y_i-\mu}{\sigma})}})  + (1-y_i)\log(1-\frac{1}{1+e^{-(\frac{y_i-\mu}{\sigma})}})\right],
\end{equation*} \label{app:logistic}
with $n$ is as the number of observations and the parameters location $\mu\in \mathbb{R}$, scale $\sigma \in \mathbb{R}^{+}$ and $x \in \mathbb{R}$.

\paragraph{Binomial Distribution}
 
The log-likelihood function for a binomial distribution is given by:

\begin{equation*}
\label{app:binomial}
    \log \left(\mathcal{L}(k|n,p) \right) = k \log(p) + (n-k) \log (1-p) + \log\left( \binom{n}{k} \right),
\end{equation*}
where $n$ is the number of trials, the parameters success probability is given by $p \in \left[0, 1 \right] $ and the number of successes is denoted as $k \in \mathbb{N}_{0}$.

\paragraph{Inverse Gamma Distribution} 
The log-likelihood function of the inverse gamma distribution is defined as:
\begin{equation*}
    \label{app:inverse-gamma}
    \log\left( \mathcal{L} (\alpha, \beta | y)\right) = - n \left( \alpha + 1 \right)\overline{\log \boldsymbol{y}} - n \log \Gamma(\alpha) + n \alpha \log \beta - \sum_{i=1}^{n} \beta y_{i}^{-1}.
\end{equation*}
with $\alpha>0$ and $\beta>0$ and where the upper bar operand indicates the arithmetic mean

\paragraph{Normal Distribution}
The log-likelihood function for a normal distribution is given by:
\begin{equation*}
    log \left(\mathcal{L}(\mu, \sigma^2 | y)\right) = -\frac{n}{2}\log(2\pi\sigma^2) - \frac{1}{2\sigma^2}\sum_{i=1}^n(y_i - \mu)^2,
\end{equation*}\label{app:normal}
where $n$ is the underlying number of observations and parameters $y \in \mathbb{R}$, location $\mu \in \mathbb{R}$ and scale $\sigma \in \mathbb{R}^{+}$.

 \paragraph{Inverse Gaussian Distribution}
 The log-likelihood function of the inverse Gaussian distribution is given by:
 \begin{equation*}
     \log \left( \mathcal{L}(\mu, \sigma |  x )\right) = \frac{n}{2} \ln(\sigma) - \sum_{i=1}^n \frac{\sigma (x_i - \mu)^2}{2 \mu^2 x_i},
 \end{equation*}\label{app:inverseGaussian}
with $n$ is as the number of observations and the parameters location $\mu\in \mathbb{R}^{+}$, scale $\sigma \in \mathbb{R}^{+}$  and $x \in \mathbb{R}^{+}$.

\paragraph{Poisson Distribution}
 The log-likelihood function for a Poisson distribution with parameter $\lambda$ is given by:

 \begin{equation*}
     \log \left(\mathcal{L}(\lambda | x)\right) = \sum_{i=1}^n [x_i \log(\lambda) - \lambda - \log(x_i!)]
     \end{equation*}\label{app:poisson}
where $x = (x_1, x_2, ..., x_n)$ is the sample, $n$ is the number of observations and $x_i$ are non-negative integers.

\paragraph{Johnson's S$_U$}
The log-likelihood function of the Johnson's S$_U$ distribution is defined as:
 $$\log\left( \mathcal{L}(\beta, \omega, \mu, \sigma | y)\right) = n \log \left[ \frac{\beta}{\omega \sqrt{2 \pi}} \right] - \frac{\beta^2}{2\omega^2} \sum_{i=1}^n \left[ \frac{(y_i - \mu)^2}{\sigma^2} + \ln \left( 1 + \frac{(y_i - \mu)^2}{\omega^2\sigma^2} \right) \right],$$
with $n$ is as the number of observations and the parameters location $\mu \in \mathbb{R}$, scale $\sigma \in \mathbb{R}^{+}$, shape $\omega \in \mathbb{R}^{+}$, skewness $\beta \in \mathbb{R}$ and $y \in \mathbb{R}$.

\paragraph{Weibull}
 The log-likelihood function of the Weibull distribution is defined as:
 $$\log\left( \mathcal{L}(\lambda, \beta, | y)\right)=n\ln \beta-n\beta\ln \lambda-\sum_{i=1}^n \left(\frac{y_i}{\lambda}\right)^\beta+(\beta-1)\sum_{i=1}^n \ln y_i,$$
 with $n$ is the number of observations and with the location $\lambda \in \mathbb{R}^{+}$, the shape $\beta \in \mathbb{R}^{+}$ and $y \in \mathbb{R}^{+}$.

\subsubsection{Deviance Measures}\label{Appendix:DevianceMeasures}
We use several deviance measures, to evaluate the model 

\paragraph{Mean Squared Error}
The mean squared error is defined as :
\begin{equation*}
    MSE = \frac{1}{n} \sum_{i=1}^n (y_i - \hat{y}_i)^2.
\end{equation*}

\paragraph{Mean Gamma Deviance}

The mean gamma deviance used for  the AirBnB dataset is defined as:
\begin{equation*}
D = \frac{2}{n} \sum_{i=1}^n  \log \left(\frac{\hat{y}_i}{y_i}\right). 
\end{equation*}

\paragraph{Area Under the Curve}
We use the Riemannian formula for the AUC. Hence the area of rectangles is defined as:

\begin{equation*}
    AR = \sum_i=1^{n-1}f(x_i)\Delta x,
\end{equation*}
and hence with larger n, the definite integral of $f$ from $a$ to $b$ is defined as:

\begin{equation*}
    \int_a^b f(x)dx = \lim_{n\to\infty} \sum_{i=0}^{n-1} f(x_i)\Delta x.
\end{equation*}

\newpage
\subsection{Further Benchmark Results}
All Benchmark results are given below. Additionally to the negative log-likelihood and the CRPS we include the presented deviance measures. As expected, mean centric models such as NAM, EBM perform well for the mean centric metrics.
Additionally, similar to figure \ref{ablation} we present the same simulation for a normal distribution:
Furthermore, we present that NAMLSS accurately captures the mean effects as well as the variance effects for the same features as used in the original NAM paper \citep{agarwal_neural_2021}. The single feature longitude and latitude effect for the California Housing dataset are accurately captured.

\begin{figure}[H]
\begin{minipage}{0.49\textwidth}
  \centering
  \includegraphics[width=0.9\textwidth]{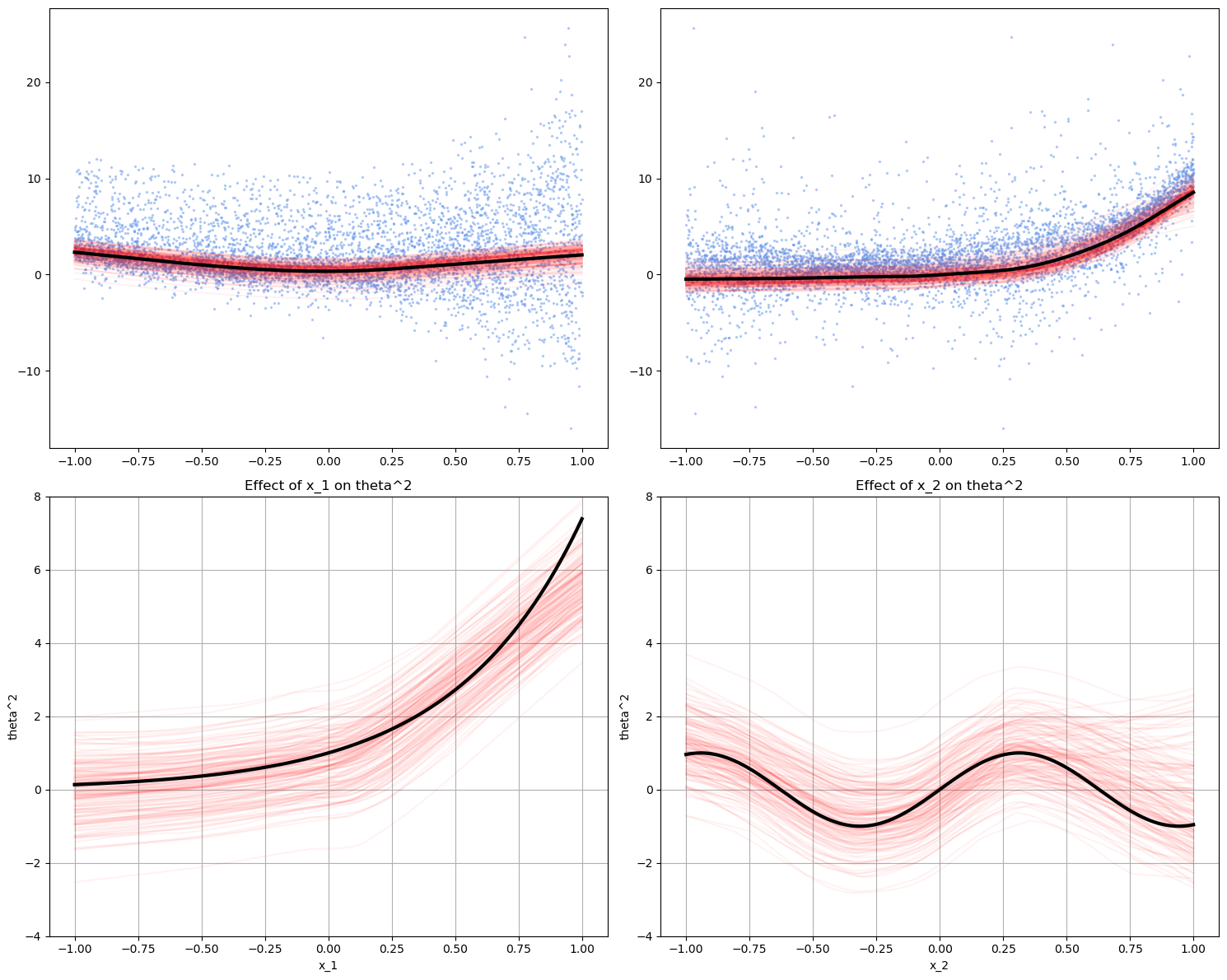}
  \caption{\small{NAMLSS distributional parameter prediction for a Normal Distribution over 150 runs. NAMLSS accurately detects the feature effects of x1 and x2 on all distributional parameters.}}
\end{minipage}\hfill
\begin{minipage}{0.49\textwidth}
  \centering
  \includegraphics[width=0.99\textwidth]{img/latitude_longitude.png}
  \caption{\small{\textbf{California Housing:} Graphs for longitude and latitude respectively learned by the NAMLSS model. NAMLSS captures changes in mean as well as variance. Therefore, the plotted standard deviations change in dependence of the longitude and latitude. The house price jumps around the location of Los Angeles are depictable. Additionally, we find a decrease in variance for areas further away from the large cities.}}
  \label{Fig:long_lat_app}
\end{minipage}
\end{figure}

All benchmark results are given below.
\newpage
\begin{table*}
\small
       \caption{Benchmark results for the FICO and the Shrutime dataset. We report continuous ranked probability score (CRPS) for the logistic distribution as there exists a close form solution as well as the AUC score.}
\begin{threeparttable}
\resizebox{\textwidth}{!}{
    \begin{tabular}{l c c c c c c }
    \toprule
     & \multicolumn{6}{c}{\textbf{Logistic}}    \\
               
     Model  & \multicolumn{3}{c}{FICO}   & \multicolumn{3}{c}{Shru Time}  \\ 
        
        & LL $\downarrow$ & AUC $\uparrow$ & CRPS $\downarrow$ & LL $\downarrow$ & AUC $\uparrow$ & CRPS $\downarrow$  \\
        \midrule
        MLP              & 1813 \footnotesize{$\pm$  (6)}            & \textbf{0.79} \footnotesize{$\pm$ (0.007)}   & 0.386  \footnotesize{$\pm$ (0.014) } & 1240 \footnotesize{$\pm$ (22) } & 0.73 \footnotesize{$\pm$ (0.011) }  & 0.296 \footnotesize{$\pm$ (0.010) }   \\
        XGBoost          & 1976 \footnotesize{$\pm$ (13)}           & 0.73 \footnotesize{$\pm$ (0.010)}    & 0.515  \footnotesize{$\pm$ (0.013) } & 1314 \footnotesize{$\pm$ (18) } & 0.79 \footnotesize{$\pm$ (0.010) }  & 0.227 \footnotesize{$\pm$ (0.011) }   \\
        NAM              & 1809 \footnotesize{$\pm$  (8)}           &0.73  \footnotesize{$\pm$ (0.010)}  & 0.372 \footnotesize{ $\pm$ (0.014)}             &  1247 \footnotesize{$\pm$ (26) } & 0.81 \footnotesize{$\pm$ (0.012) }  & 0.231 \footnotesize{$\pm$ (0.007) }    \\
        EBM              & 1944 \footnotesize{$\pm$ (21)}           & 0.73  \footnotesize{$\pm$ (0.010)}  & 0.511 \footnotesize{ $\pm$ (0.013)}            &  1290 \footnotesize{$\pm$ (26) }  & 0.83 \footnotesize{$\pm$ (0.012) }  & 0.201 \footnotesize{$\pm$ (0.016) }    \\
        NodeGAM          & 1942 \footnotesize{$\pm$ (21)}           & 0.72 \footnotesize{$\pm$ (0.006)} & 0.513  \footnotesize{$\pm$ (0.027) }             &  1308 \footnotesize{$\pm$ (29) } & 0.81 \footnotesize{$\pm$ (0.012) }  &  0.205 \footnotesize{$\pm$ (0.015) }    \\
        GAMLSS          & 1321  \footnotesize{$\pm$ (30)}           & 0.78 \footnotesize{$\pm$ (0.009)}        & 0.392  \footnotesize{$\pm$ (0.005) }              &   391 \footnotesize{$\pm$ (126) } & 0.77 \footnotesize{$\pm$ (0.014) }  & 0.119 \footnotesize{$\pm$ (0.006) }    \\
        gamboostLSS      & 1191 \footnotesize{$\pm$ (30)}           & \textit{0.79} \footnotesize{$\pm$ (0.008)}      & 0.370  \footnotesize{$\pm$ (0.007) }         &   $^*$ & $^*$  & $^*$   \\
        \hline
        DDNN             & 1230 \footnotesize{$\pm$ (48)}          & 0.73 \footnotesize{$\pm$ (0.002)} & \textit{0.342}  \footnotesize{$\pm$ (0.012)}      &  -211 \footnotesize{$\pm$ (364) } & 0.81 \footnotesize{$\pm$ (0.011) }  & 0.145 \footnotesize{$\pm$ (0.012) }    \\
        NAMLSS\tnote{1}  & 1201 \footnotesize{$\pm$ (41)}          & 0.73 \footnotesize{$\pm$ (0.010)} & 0.347  \footnotesize{$\pm$ (0.019)}               &  \textit{-220} \footnotesize{$\pm$ (210) } & \textbf{0.85} \footnotesize{$\pm$ (0.040) }  &  \textit{0.111} \footnotesize{$\pm$ (0.015) }     \\
        NAMLSS\tnote{2}  & \textbf{1160} \footnotesize{$\pm$ (49)} & 0.72 \footnotesize{$\pm$ (0.008)} & \textbf{0.328}  \footnotesize{$\pm$ (0.013) }     &  \textbf{-237} \footnotesize{$\pm$ (219) } & \textit{0.84} \footnotesize{$\pm$ (0.020) }  & \textbf{0.107} \footnotesize{$\pm$ (0.003) }     \\
        \bottomrule   
    \end{tabular}}
    \begin{tablenotes}
      \small
      \begin{enumerate*}
      \item[$^{*}$] gamboostLSS was not able to execute.
      \end{enumerate*}
    \end{tablenotes}
    \end{threeparttable}
\end{table*}

\begin{table*}
\small
       \caption{Benchmark results for the California Housing and the Abalone dataset. We report continuous ranked probability score (CRPS) as there exists a close form solution. For metrics detecting the accuracy of point predictions, we find the models that specifically minimize the MSE to excel.}
\begin{threeparttable}
\resizebox{\textwidth}{!}{
    \begin{tabular}{l c c c c c c}
    \toprule
               
         Model  &  \multicolumn{3}{c}{CA Housing} & \multicolumn{3}{c}{Abalone}   \\ 
        
        & LL$\downarrow$ & MSE $\downarrow$ & CRPS $\downarrow$& LL $\downarrow$ & MSE $\downarrow$ & CRPS $\downarrow$   \\
        \midrule
        MLP              & 4191 \footnotesize{$\pm$ (42)} & \textbf{0.197} $\pm$ (0.005) & 0.264 $\pm$ (0.004) & 966.2  \footnotesize{$\pm$ (27)} & 0.475 $\pm$ (0.044) & 0.472 $\pm$ (0.013) \\     
        XGBoost          & 4219 \footnotesize{$\pm$ (40)} & 0.211 $\pm$(0.005)  & 0.271 $\pm$ (0.002) & 982.0  \footnotesize{$\pm$ (33)} & 0.515 $\pm$ (0.028) & 0.504 $\pm$ (0.015) \\    
        NAM              & 4251 \footnotesize{$\pm$ (43)} & 0.273 $\pm$(0.037)  & 0.370 $\pm$ (0.045) & 956.8   \footnotesize{$\pm$ (22)}& \textit{0.454} $\pm$ (0.024) & 0.484 $\pm$ (0.015)    \\  
        EBM              & 4202 \footnotesize{$\pm$ (42)} & 0.203 $\pm$(0.004)  & 0.297 $\pm$ (0.002) & 965.1  \footnotesize{$\pm$ (22)} & 0.474 $\pm$ (0.025) & 0.491 $\pm$ (0.012)     \\
        NodeGAM          & 4206 \footnotesize{$\pm$ (89)} & 0.242 $\pm$ (0.007) & 0.350 $\pm$ (0.004) & 958.3  \footnotesize{$\pm$ (23)} & 0.461$\pm$ (0.033) & 0.486 $\pm$ (0.017)     \\
        GAMLSS           & 3512 \footnotesize{$\pm$ (67)} & 0.398 \footnotesize{$\pm$ (0.040)} & 0.315 \footnotesize{$\pm$ (0.005) }  &  870.8 \footnotesize{$\pm$ (16) } & 0.497 \footnotesize{$\pm$ (0.032) } & 0.353 \footnotesize{ $\pm$ (0.007) }  \\
        gamboostLSS      & 3812 \footnotesize{$\pm$ (52)} &  0.415 \footnotesize{$\pm$ (0.024)} & 0.336  \footnotesize{$\pm$ ( 0.004) } &  815.1 \footnotesize{$\pm$ (29) } & 0.524 \footnotesize{$\pm$ (0.039) }  & 0.370 \footnotesize{ $\pm$ (0.011) } \\
        \hline
        DDNN             & 2681 \footnotesize{$\pm$ (1279)} & \textit{0.197} $\pm$ (0.005) & \textbf{0.193} $\pm$ (0.002) & 897.2 \footnotesize{$\pm$ (159)} & \textbf{0.444} $\pm$ (0.026) & \textbf{0.338} $\pm$ (0.009)       \\
        NAMLSS\tnote{1}  &  \textit{2667} \footnotesize{$\pm$ (91)}  & 0.245 $\pm$ (0.004) & 0.287 $\pm$ (0.019) & \textit{869.8} \footnotesize{$\pm$ (118)}             & 0.496 $\pm$ (0.042) & 0.357 $\pm$ (0.013)        \\  
        NAMLSS\tnote{2}  &  \textbf{2329} \footnotesize{$\pm$ (176)} & 0.265 $\pm$ (0.005) & \textit{0.264} $\pm$ (0.009) & \textbf{802.3} \footnotesize{$\pm$ (41)}     & 0.486$\pm$ (0.043) & \textit{0.353} $\pm$ (0.014)        \\
        \bottomrule
    \end{tabular}}
    \end{threeparttable}
    
\end{table*}

\begin{table*}
\small
\caption{Benchmark results for the Insurance dataset and munich dataset. }
\begin{threeparttable}
\resizebox{\textwidth}{!}{
    \begin{tabular}{l c c c c c c}
    \toprule
     &  \multicolumn{6}{c}{\textbf{Inv. Gamma}}   \\
               
     Model  & \multicolumn{3}{c}{Melbourne}   & \multicolumn{3}{c}{Munich }  \\
        
        &  LL $\downarrow$ & Gamma dev. $\downarrow$ & KL-divergence $\downarrow$ &  LL $\downarrow$ & Gamma dev. $\downarrow$ & KL-divergence $\downarrow$ \\
        \midrule
        MLP                        & 22999 \footnotesize{$\pm$ (232)} & 1.090 \footnotesize{$\pm$ (0.52)} & 5.23 \footnotesize{$\pm$ (1.09)}          & 6827 \footnotesize{$\pm$ (178)}             & 0.55 \footnotesize{$\pm$ (0.04)} & 0.119    \footnotesize{$\pm$ (0.034)} \\     
        XGBoost                    & 20471 \footnotesize{$\pm$ (242)} & 1.088 \footnotesize{$\pm$ (0.498)} & 5.23 \footnotesize{$\pm$ (1.17)}           &  5618 \footnotesize{$\pm$ (152)}            & \textbf{0.48} \footnotesize{$\pm$ (0.09)} & 0.125 \footnotesize{$\pm$ (0.015)} \\     
        NAM                        & 25375\footnotesize{$\pm$ (844)} & 0.871\footnotesize{$\pm$ (0.209)} & 4.70\footnotesize{$\pm$ (1.93)}           &  5892 \footnotesize{$\pm$ (37)}             & 0.72 \footnotesize{$\pm$ (0.10)} & 0.169    \footnotesize{$\pm$ (0.048)}    \\    
        EBM                        & 20361\footnotesize{$\pm$ (207)} & 1.206\footnotesize{$\pm$ (0.651)} & 5.59 \footnotesize{$\pm$ (0.64)}           &  5474 \footnotesize{$\pm$ (56)}             & \textit{0.49} \footnotesize{$\pm$ (0.09) } & 0.117  \footnotesize{$\pm$ (0.021)} \\
        NodeGAM                    & 20790\footnotesize{$\pm$ (29)} & 1.110 \footnotesize{$\pm$ (0.466)} & 5.66\footnotesize{$\pm$ (0.59)}           &  5984 \footnotesize{$\pm$ (135)}            & 0.57 \footnotesize{$\pm$ (0.05) } & 0.112 \footnotesize{$\pm$ (0.059)}  \\
        GAMLSS          & 26353 \footnotesize{$\pm$ (45)}           & \textbf{0.737} \footnotesize{$\pm$ (0.069)} & 4.30 \footnotesize{$\pm$ (0.01)}             &  5419 \footnotesize{$\pm$ (61) } & 0.53 \footnotesize{$\pm$ (0.06) }  & 0.148 \footnotesize{$\pm$ (0.056) }  \\
        gamboostLSS          & 26436 \footnotesize{$\pm$ (48)}           & \textit{0.761} \footnotesize{$\pm$ (0.044)} & 4.29 \footnotesize{$\pm$ (0.02)}         &   5421\footnotesize{$\pm$ (33) } & 0.58 \footnotesize{$\pm$ (0.10) }  & 0.148 \footnotesize{$\pm$ (0.056) }  \\
        \hline
        DDNN                       & 26353\footnotesize{$\pm$ (45)} & 0.987\footnotesize{$\pm$ (0.304)} & 4.21\footnotesize{$\pm$ (2.02)}           & 5555 \footnotesize{$\pm$ (34)}              & 0.69 \footnotesize{$\pm$ (0.04)} & \textit{0.011}  \footnotesize{$\pm$ (0.015)}       \\
        NAMLSS\tnote{1}            & \textbf{19517}\footnotesize{$\pm$ (68)} & 0.962\footnotesize{$\pm$ (0.272)} & \textit{4.23}\footnotesize{$\pm$ (2.09)}           & \textbf{5383} \footnotesize{$\pm$ (24)}     & 0.59 \footnotesize{$\pm$ (0.09)} & 0.014    \footnotesize{$\pm$ (0.018)}             \\  
        NAMLSS\tnote{2}            & \textit{19675}\footnotesize{$\pm$ (67)} & 0.966\footnotesize{$\pm$ (0.285)} & \textbf{4.18}\footnotesize{$\pm$ (2.06)}           &  \textit{5422} \footnotesize{$\pm$ (22)}    & 0.59 \footnotesize{$\pm$ (0.10)} & \textbf{0.011} \footnotesize{$\pm$ (0.015)}        \\
        \bottomrule
    \end{tabular}}
    \end{threeparttable}
    
\end{table*}

\begin{table*}
\small
       \caption{Benchmark results for the Telco and the -. We report continuous ranked probability score (CRPS) for the logistic distribution as there exists a close form solution as well as the AUC score.}
\begin{threeparttable}
\resizebox{\textwidth}{!}{
    \begin{tabular}{l c c c c c c }
    \toprule
     & \multicolumn{3}{c}{\textbf{Logistic}} & \multicolumn{3}{c}{\textbf{Normal}}    \\
               
     Model  & \multicolumn{3}{c}{Telco}   & \multicolumn{3}{c}{Insurance}  \\ 
        
        & LL $\downarrow$ & AUC $\uparrow$ & CRPS $\downarrow$ & LL $\downarrow$ & MSE $\downarrow$ & CRPS $\downarrow$  \\
        \midrule
        MLP              & 1027 \footnotesize{$\pm$ (19)} & 0.69 \footnotesize{$\pm$ (0.012)}  & 0.307 \footnotesize{$\pm$ (0.012)} & 266.8 \footnotesize{$\pm$ (11)}             & \textit{0.17} $\pm$ (0.027) & 0.276 $\pm$ (0.019)     \\
        XGBoost          & 1123 \footnotesize{$\pm$ (22)} & 0.73 \footnotesize{$\pm$ (0.005)}  & 0.331 \footnotesize{$\pm$ (0.014)} & 266.8 \footnotesize{$\pm$ (9)}              & 0.19 $\pm$(0.024)  & 0.248 $\pm$ (0.012)      \\
        NAM              & 1023 \footnotesize{$\pm$ (28)} & 0.76 \footnotesize{$\pm$ (0.011)}  & 0.279 \footnotesize{$\pm$ (0.008)} & 474.4 \footnotesize{$\pm$ (73)}             & 0.26 $\pm$(0.031)  & 0.381 $\pm$ (0.022)   \\
        EBM              & 1094 \footnotesize{$\pm$ (22)} & \textit{0.76} \footnotesize{$\pm$ (0.007)}  & 0.304 \footnotesize{$\pm$ (0.012)} & 263.8 \footnotesize{$\pm$ (10)}             & \textbf{0.14} $\pm$(0.018)  & \textbf{0.204} $\pm$ (0.009)   \\
        NodeGAM          & 1097 \footnotesize{$\pm$ (27)} & 0.76 \footnotesize{$\pm$ (0.020)}  & 0.303 \footnotesize{$\pm$ (0.019)} & 279.1 \footnotesize{$\pm$ (11)}             & 0.26 $\pm$ (0.031) & 0.366 $\pm$ (0.026)   \\
        GAMLSS          &   85\footnotesize{$\pm$ (173)}           & \textbf{0.83} \footnotesize{$\pm$ (0.011)} & 3.38 \footnotesize{$\pm$ (6.369)}              &   175.5 \footnotesize{$\pm$ (28)} & 0.241 \footnotesize{$\pm$ (0.033) }  & 0.263 \footnotesize{$\pm$ (0.020) }  \\
        gamboostLSS      &  $^*$            & $^*$  & $^*$         &   173.0 \footnotesize{$\pm$ (28) } & 0.268 \footnotesize{$\pm$ (0.044) }  & 0.248 \footnotesize{$\pm$ (0.019) }     \\
        \hline
        DDNN             & 27 \footnotesize{$\pm$ (314)}  & 0.73 \footnotesize{$\pm$ (0.015)}  & 0.188 \footnotesize{$\pm$ (0.015)} & 178.2 \footnotesize{$\pm$ (30)}             & 0.17 $\pm$ (0.028) & \textit{0.206} $\pm$ (0.016)   \\
        NAMLSS\tnote{1}  & \textbf{-22} \footnotesize{$\pm$ (137)} & 0.64 \footnotesize{$\pm$ (0.018)}  & \textbf{0.133} \footnotesize{$\pm$ (0.013)} &  \textit{172.7} \footnotesize{$\pm$ (23)}   & 0.27 $\pm$ (0.044) & 0.257 $\pm$ (0.020)     \\
        NAMLSS\tnote{2}  & \textit{-11} \footnotesize{$\pm$ (114)} & 0.65 \footnotesize{$\pm$ (0.017)}  & \textit{0.135} \footnotesize{$\pm$ (0.008)} &  \textbf{172.6} \footnotesize{$\pm$ (20)}   & 0.27 $\pm$ (0.049) & 0.257 $\pm$ (0.022)  \\
        \bottomrule   
    \end{tabular}}
     \begin{tablenotes}
      \small
      \begin{enumerate*}
      \item[$^{*}$] gamboostLSS was not able to execute.
      \end{enumerate*}
    \end{tablenotes}
    \end{threeparttable}
\end{table*}

\subsubsection{Synthetic Benchmarks}\label{app:reproducability}
The benchmark study for used real-world datasets was performed under similar conditions. All datasets are publicly available and we describe every preprocessing step as well as all model specifications in detail in the following. 

\paragraph{Synthetic Data Generation}
For the simulation of the data, respectively their underlying distribution parameters $\theta=\left( \theta^{(1)},\theta^{(2)},\theta^{(3)},\theta^{(4)} \right)$, the following assumptions are made:
\begin{align*}
    \theta^{(1)} &= \frac{30}{13} x_1  \left( (3x_2 + 1.5) - 2   \sin \left( \frac{x_3}{2} \right) \right)^{-1} + \frac{113}{115}  x_4 + 0.1  x_5, \\
    \theta^{(2)} &= \exp \left( -0.0035  x_1 + ( x_2 - 0.23)^2 - 1.42  x_3 \right) + 0.0001 x_4, \\
    \theta^{(3)} &= \frac{1}{42} (4 x_1 - 90 x_2), \\
    \theta^{(4)} &=  exp\left(0.0323 * x_2 + 0.0123 - 0.0234 * x_4\right),
\end{align*}\label{app:daten_simulation_formel}

where each of the five input vectors $x_j$ is sampled from a uniform distribution $\mathcal{U}(0,1)$, with a total of $n=3000$ observations per data set.

\newpage
\subsection{Activation Functions}

For DDNN and NAMLSS, independent of the implementation, we use a Softplus activation for the scale parameter $\sigma^2$ to ensure non-negativity and a linear activation for the mean $\mu$.

For the AirBnB datasets, also analyzed by \citet{rugamer2020semi}, we assume an Inverse Gamma distribution $\mathcal{IG}(\alpha, \beta)$ as the underlying data distribution (see equation (\ref{app:inverse-gamma}) for the log-likelihood). 
For NAMLSS as well as DDNN we have to adjust the activation functions, as both models minimize the log-likelihood via the parameters $\alpha$ and $\beta$. 
However, the mean prediction resulting from these parameters is defined via: $$\mu = \frac{\beta}{\alpha -1}$$ and is hence only defined for $\alpha > 1$. 
The activation functions thus need to ensure an $\alpha$ prediction that is larger than 1 and a $\beta$ prediction that is larger than 0. 
Hence we again use a Softplus activation for the $\beta$ output layer \footnote{Interestingly, the NAM did not converge using the Softplus activation function as the MLP did. 
Using the Softplus activation resulted in tremendously large mean gamma deviances and log-likelihoods, as the model kept predicting values that were nearly zero. 
Hence, we were only able to achieve good results for the NAM using the activation function given by formula (\ref{eq:gamma_activation}).}. 
For the $\alpha$ prediction, we use the following activation function element-wise:

\begin{equation}
    \label{eq:gamma_activation}
    h(x) = \begin{cases} \log(1+\exp(x)), & \text{if } \log(1+\exp(x)) > 1, \\
    \frac{1}{\log(1+\exp(x))}, & \text{else}.
    \end{cases}
\end{equation}

To compute the log-likelihood for the models resulting in a mean prediction we compute the parameters $\alpha$ and $\beta$ as follows:

$$\alpha = \frac{\mu^2}{\sigma^2 + 2},$$
$$\beta = \mu \frac{\mu^2}{\sigma^2+1},$$  

with $\sigma^2$ denoting the variance of the mean predictions. 
For XGBoost and EBM we use a simple transformation of the target variable to ensure that $\mu > 0$. 
Hence we fit the model on $\log(y)$ and re-transform the predictions accordingly with $\exp(\hat{y})$. 

For a (binary) classification benchmark we use the FICO dataset \citep{FICO}, the Shrutime dataset and the Telco dataset. 
A logistic distribution, $\mathcal{LO}\left(\mu, s \right)$, of the underlying response variable was assumed (see equation (\ref{app:logistic}) for the log-likelihood). 
Again, we use the true standard deviation of the underlying data for the models only resulting in a mean prediction. The models resulting in a mean prediction use binary cross-entropy as the loss function and hence a sigmoid activation function on the output layer.

\subsubsection{Preprocessing}
We implement the same preprocessing for all used datasets and only slightly adapt the preprocessing of the target variable for the two regression problems, California Housing and Insurance.
We closely follow \citet{gorishniy_revisiting_2021} in their preprocessing steps and use the preprocessing also implemented by \citet{agarwal_neural_2021}. Hence all numerical variables are scaled between -1 and 1, all categorical features are one-hot encoded. In contrast to \citet{gorishniy_revisiting_2021} we do not implement quantile smoothing, as one of the biggest advantages of neural models is the capability to model jagged shape functions.
We use 5-fold cross-validation and report mean results as well as the standard deviations over all datasets. For reproducibility, we use the sklearn \citep{pedregosa_scikit-learn_2011} Kfold function with a random state of 101 and shuffle equal to true for all datasets.
For the two regression datasets, we implement a standard normal transformation of the target variable. This results in better performances in terms of log-likelihood for all models only predicting a mean and is hence even disadvantageous for the presented NAMLSS framework.

\subsubsection{Datasets}

\paragraph{California Housing}

\begin{table*}[h]
\centering
\small
       \caption{Statistics of the benchmarking datasets.}
\begin{threeparttable}
    \begin{tabular}{l|l l l l }
        Dataset            & No. Samples & No. Features & Distribution                                 & Task \\
        \hline\hline
        California Housing & $20640$     & 8            & Normal $\mathcal{N}(\mu, \sigma)$            & Regression \\
        Insurance          & $1338$      & 6            & Normal $\mathcal{N}(\mu, \sigma)$            & Regression \\
        Abalone            & $4177$      & 10           & Normal $\mathcal{N}(\mu, \sigma)$            & Regression \\
        Munich             & $4568$      & 9            & Inverse Gamma $\mathcal{IG}(\alpha, \beta)$  & Regression \\
        Melbourne          & $16868$     & 11           & Inverse Gamma $\mathcal{IG}(\alpha, \beta)$  & Regression \\
        Fico               & $10459$     & 23           & Logistic $\mathcal{LO}(\mu, s)$              & Classification \\
        Shrutime           & $10000$     & 10           & Logistic $\mathcal{LO}(\mu, s)$              & Classification \\
        Telco              & $7032$      & 19           & Logistic $\mathcal{LO}(\mu, s)$              & Classification \\
        \bottomrule
    \end{tabular}
    \end{threeparttable}
\end{table*}

The California Housing (CA Housing) dataset \cite{pace1997sparse} is a popular publicly available dataset and was obtained from sklearn \cite{pedregosa_scikit-learn_2011}. It is also used as a benchmark in \cite{agarwal_neural_2021} and \citet{gorishniy_revisiting_2021} and we achieve similar results concerning the MSE for the models which were used in both publications.
The dataset contains the house prices for California homes from the U.S. census in 1990. The dataset is comprised of 20640 observations and besides the logarithmic median house price of the blockwise areas as the target variable contains eight predictors. As described above, we additionally standard normalize the target variable. All other variables are preprocessed as described above.

\paragraph{Insurance}
The Insurance dataset is another regression type dataset for predicting billed medical expenses \citep{lantz2019machine}. The dataset is publicly available in the book \textit{Machine Learning with R} by \citet{lantz2019machine}. Additionally, the data is freely available on  Github (\url{https://github.com/stedy/Machine-Learning-with-R-datasets}) and Kaggle (\url{https://www.kaggle.com/code/gloriousc/insurance-forecast-by-using-linear-regression/data}).
It is a small dataset with only 1338 observations. The target variable is \textit{charges}, which represents the \textit{Individual medical costs billed by health insurance}. Similar to the California Housing regression we standard normalize the response. 
Additionally, the dataset includes 6 feature variables. They are preprocessed as described above, which, due to one-hot encoding leads to a feature matrix with 9 columns.

\paragraph{Abalone}
The Abalone dataset contains information for the prediction of the age of abalone, a type of sea snail, based on their physical measurements. The data set is taken from the original publication \citep{nash1994population} and today is a part of the UCI Machine Learning Repository (\url{https://archive.ics.uci.edu/ml/datasets/abalone}).
A dataset of 4177 observations, 10 features and one response variable is obtained after processing the data.

\paragraph{Munich}
For the AirBnB data, we orientate on \citet{rugamer2020semi} and used the data for the city of Munich. 
The dataset is also publicly available and was taken from Inside AirBnB (\url{http://insideairbnb.com/get-the-data/}) on January 15, 2023. 
After excluding the variables \textit{ID}, \textit{Name}, \textit{Host ID}, \textit{Host Name}, \textit{Last Review} and after removing rows with missing values the dataset contains 4568 observations.
Additionally, we drop the \textit{Neighbourhood} variable as firstly the predictive power of that variable is limited at best and secondly not to create too large feature matrices for GAMLSS. 
Hence, in addition to the target variable, the dataset contains 9 variables. 
All preprocessing steps are subsequently performed as described above and the target variable, \textit{Price}, is not preprocessed at all.

\paragraph{Melbourne}
The dataset is also publicly available and was taken from Inside AirBnB (\url{http://insideairbnb.com/get-the-data/}).
The second Airbnb dataset (Melbourne) originates from the same source as the Munich Airbnb dataset. The data processing follows the same procedure as described in the Munich section. All preprocessing steps are then performed as described above and the target variable \textit{Price} is not preprocessed at all.

\paragraph{FICO}
Similar to \citet{agarwal_neural_2021} we also use the FICO dataset for our benchmarking study. However, we use it as described on the website and hence use the \textit{Risk Performance} as the target variable.
A detailed description of the features and their meaning is available at the Explainable Machine Learning Challenge (\url{https://community.fico.com/s/explainable-machine-learning-challenge}). The dataset is comprised of 10459 observations. We did not implement any preprocessing steps for the target variable.

\paragraph{Shrutime}
This dataset contains information on the customers of a bank and the target variable is a binary variable reflecting whether the customer has left the bank (closed his account) or remains a customer. The corresponding data set can be found at
Kaggle (\url{https://www.kaggle.com/datasets/shrutimechlearn/churn-modelling}) and is introduced by \citet{churn2019data}.
After the processing described above, the set consists of $10000$ observations, each with $10$ features.

\paragraph{Telco}
The Telco customer churn data contains information about a fictitious telco company that provided home phone and internet services to 7043 customers. It details which customers left, stayed or signed up for their service. Several key demographics are included for each customer, as well as a satisfaction score, a churn score and a customer lifetime value (CLTV) index and was introduced by \citet{ibm2019teclo}.
After the processing described above, the set consists of $7043$ observations, each with $19$ features.

\subsubsection{Model Architectures \& Hyperparameters}

As we do not implement extensive hyperparameter tuning for the presented NAMLSS framework, we do not perform hyperparameter tuning for the comparison models. We fit all models without an intercept.
However, we try to achieve the highest comparability by choosing similar modelling frameworks, network architectures and hyperparameters where possible. 
All neural models are hence fit with identical learning rates, batch sizes, hidden layer sizes, activation functions and regularization techniques.
Through all neural models and all datasets, we use the ADAM optimizer \citep{kingma2014adam} with a starting learning rate of 1e-04. 
For the larger datasets, \textit{California Housing}, \textit{Abalone}, \textit{FICO}, \textit{Telco} and \textit{Shrutime} we orient on \citet{agarwal_neural_2021} and use larger batch sizes of 1024. 
For the smaller dataset, \textit{Insurance}, we use a smaller batch size of 256 and for the \textit{Munich} and \textit{Melbourne} dataset we use a batch size of 512. 
For every dataset and for every neural model, the maximum number of epochs is set to 2000. 
However, we implement early stopping with a patience of 150 epochs and no model over no fold and no dataset ever trained for the full 2000 epochs. 
Additionally, we reduce the learning rate with a factor of 0.95 with patience of 10 epochs for all models for all datasets. 
We use the Rectified Linear Unit (ReLU) activation function for all hidden layers for all models:

\begin{equation*}
    h(x) = \begin{cases} 0, & x < 0 \\
    x, &\text{else}.
    \end{cases}
\end{equation*}\label{eq:relu} 
We also experimented with the Exponential centred hidden Unit (ExU) activation function presented by \citet{agarwal_neural_2021} but found no improvement in model performance and even a slight deterioration for most models.

For the statistical models used from the GAMLSS and gamboostLSS frameworks, we do not optimize the model hyperparameters, as with neural networks. 
We use the respective default settings unless otherwise stated in the modelling descriptions included in the Appendix. 
We try to keep the model settings equal between all models, if applicable. 
All GAMLSS models use the same RS solver proposed by \cite{rigby2005generalized}, in cases where this approach does not lead to convergence, the alternative CG solver presented by \cite{cole1992smoothing} is employed. 
To exclude possible numerical differences, the same distributions from the GAMLSS R package are used for modelling the response distribution and calculating the log-likelihoods.
gamboostLSS allows the use of different boosting approaches. Here we use the implemented boosting methods based on GAMs and GLMs and choose the model that performs better in terms of log-likelihood and the assumed loss. 

\paragraph{California Housing and Abalone}

\begin{table}[h]
\centering
\small
       \caption{Hyperparameters for the neural models for the California Housing and the Abalone dataset}
\begin{threeparttable}
    \begin{tabular}{l|l l l l l }
        Hyperparameter      & NAMLSS$^1$ & NAMLSS$^2$ &  DNN & MLP & NAM  \\
        \hline\hline
        Learning Rate       & 1e-04                         & 1e-04                         & 1e-04                         & 1e-04                      & 1e-04                      \\
        Dropout             & 0.25                          & 0.25                          & 0.25                          & 0.25                       & 0.25                       \\
        \multirow{2}{*}{Hidden Layers}       & $[1000, 500, $    & $[1000, 500,$    & $[1000, 500,$    & $[1000, 500,$ & $[1000, 500, $ \\
        & $100, 50, 25]$    & $100, 50, 25]$    & $100, 50, 25]$    & $100, 50, 25]$ & $100, 50, 25]$ \\
        LR Decay, Patience  & 0.95 - 10                     & 0.95 - 10                     & 0.95 - 10                     & 0.95 - 10                  & 0.95 - 10                  \\
        Activation          & ReLU                          & ReLU                          & ReLU                          & ReLU                       & ReLU                       \\
        Output Activation   & Linear, Softplus              & Linear, Softplus              & Linear, Softplus              & Linear                     & Linear                     \\
        \bottomrule
    \end{tabular}
    \begin{tablenotes}
      \small
      \begin{enumerate*}
      \item[$^{1}$] With 2 $\times 8 $ subnetworks. See Table \ref{fig:namlss_structure1} for an exemplary network structure.\\ 
      \item[$^{2}$] With 8 subnetworks and each subnetwork returning a parameter for the location and shape respectively. See Table \ref{fig:namlss_structure2} for an exemplary network structure.\\ 
      \end{enumerate*}
      
    \end{tablenotes}
    \end{threeparttable}
\end{table}

We orient again on \citet{agarwal_neural_2021} and use the following hidden layer sizes for all networks: $[1000, 500, 100, 50, 25]$. 
The second hidden layer is followed by a 0.25 dropout layer. 
While subsequently the NAM and NAMLSS have much more trainable parameters than the MLP and the DNN, we find that the MLP and DNN outperform the NAM and NAMLSS in terms of mean prediction. 
Additionally, we encountered severe overfitting when using the same number of parameters in an MLP as in the NAM and NAMLSS implementation. 
For the mean predicting models, we use a one-dimensional output layer with a linear activation. 
For the DNN and both NAMLSS implementations, we use a linear activation over the mean prediction and a Softplus activation for the variance prediction with:

$$h(x) = \log(1+\exp(x)).$$

For the NAMLSS implementation depicted in Figure \ref{fig:namlss_structure1} we use a smaller network structure for predicting the variance with two hidden layers of sizes 50 and 25 without any form of regularization as \citet{durr2020probabilistic} found that using smaller networks for predicting the scale parameters is sufficient. 
For XGBoost we use the default parameters from the Python implementation. For the Explainable Boosting machines, we increased the number of maximum epochs to the default value of 5000 but set the early stopping patience considerably lower to 10, as otherwise, the model reached far worse results compared to the other models. 
We additionally increased the learning rate to 0.005 compared to the learning rate used in the neural approaches as a too small learning rate resulted in bad results. 
Otherwise, we kept all other hyperparameters as the default values.
The GAMLSS and gamboostLSS models assume a normal distribution, with a location estimator $\mu$ employing an identity link and a scale estimator $\sigma$ with a $\log$-link function.
Due to numerical instabilities, we choose to use the GLM-based boosting method instead of the default GAM-based version.

\paragraph{Insurance}
\begin{table}[h]
\centering
\small
       \caption{Hyperparameters for the neural models for the Insurance dataset}
\begin{threeparttable}
    \begin{tabular}{l|l l l l l}
        Hyperparameter      & NAMLSS$^1$ & NAMLSS$^2$ &  DNN & MLP & NAM  \\
        \hline\hline
        Learning Rate       & 1e-04                 & 1e-04                 & 1e-04                 & 1e-04             & 1e-04                \\
        Dropout             & 0.5                   & 0.5                   & 0.5                   & 0.5               & 0.5                 \\
        Hidden Layers       & $[250, 50, 25]$       & $[250, 50, 25]$       & $[250, 50, 25]$       & $[250, 50, 25]$   & $[250, 50, 25]$     \\
        LR Decay, Patience  & 0.95 - 10             & 0.95 - 10             & 0.95 - 10             & 0.95 - 10         & 0.95 - 10            \\
        Activation          & ReLU                  & ReLU                  & ReLU                  & ReLU              & ReLU                 \\
        Output Activation   & Linear, Softplus      & Linear, Softplus      & Linear, Softplus      & Linear            & Linear               \\
        \bottomrule
    \end{tabular}
    \begin{tablenotes}
      \small
      \begin{enumerate*}
      \item[$^{1}$] With 2 $\times 9 $ subnetworks. See Table \ref{fig:namlss_structure1} for an exemplary network structure.\\ 
      \item[$^{2}$] With 9 subnetworks and each subnetwork returning a parameter for the location and shape respectively. See Table \ref{fig:namlss_structure2} for an exemplary network structure.\\ 
      \end{enumerate*}
      
    \end{tablenotes}
    \end{threeparttable}
\end{table}
As the insurance dataset is considerably smaller than all other datasets we use slightly different model structures, as the model structure used for the California Housing and Abalone datasets led to worse results.
Hence, for all neural models, we use hidden layers of sizes $[250, 50, 25]$. The first layer is followed by a 0.5 dropout layer. 
Again, we use a simple linear activation for the models only predicting the mean and a linear and a Softplus activation for the models predicting the mean and the variance respectively. 
For the first NAMLSS implementation (see Figure \ref{fig:namlss_structure1}) we again use a smaller network for predicting the variance with just one hidden layer with 50 neurons.

For XGBoost and EBM we use the same hyperparameter specifications as for the California Housing and Abalone datasets.

The GAMLSS and gamboostLSS models assume a normal distribution, with a location estimator $\mu$ employing an identity link and a scale estimator $\sigma$ with a $\log$-link function.
The boosting for location, scale and shape method employed uses the GLM based, instead of the GAM, based version.

\paragraph{FICO, Telco and Shrutime}
For the logistic datasets, we use the exact same model structure as for the Insurance dataset, as the model structures implemented for the California Housing dataset resulted in worse results. However, as it is a binary classification problem we use a Sigmoid activation for the MLP as well as the NAM. For the DNN and both NAMLSS implementations, we use a Sigmoid activation for the location and a Softplus activation for the scale. To generate the log-likelihoods for the models only predicting a mean, we again use the true standard deviation of the underlying data.

For XGBoost and EBM we had to adjust the hyperparameters in order to get results comparable to the MLP, NAM or NAMLSS.
Hence, for EBM we use 10 as the maximum number of leaves, 100 early stopping rounds and again the same learning rate of 0.005.

For XGboost we use 500 estimators with a maximum depth of 15. $\eta$ is set to 0.05.

For the GAMLSS and gamboost models we use a logistic distribution to model the response distribution, where $\mu$ estimator uses identity and the $\sigma$ estimator uses a log-link function.

\begin{table}[H]
\centering
\small
       \caption{Hyperparameters for the neural models for the FICO, Telco and Shrutime datasets}
\begin{threeparttable}
    \begin{tabular}{l|l l l l l}
        Hyperparameter      & NAMLSS$^1$ & NAMLSS$^2$ &  DNN & MLP & NAM \\
        \hline\hline
        Learning Rate       & 1e-04                 & 1e-04                 & 1e-04                 & 1e-04           & 1e-04                \\
        Dropout             & 0.5                   & 0.5                   & 0.5                   & 0.5               & 0.5                 \\
        Hidden Layers       & $[250, 50, 25]$       & $[250, 50, 25]$       & $[250, 50, 25]$       & $[250, 50, 25]$ & $[250, 50, 25]$      \\
        LR Decay, Patience  & 0.95 - 10             & 0.95 - 10             & 0.95 - 10             & 0.95 - 10       & 0.95 - 10             \\
        Activation          & ReLU                  & ReLU                  & ReLU                  & ReLU            & ReLU                  \\
        Output Activation   & Sigmoid, Softplus      & Sigmoid, Softplus      & Sigmoid, Softplus   & Sigmoid         & Sigmoid               \\
        \bottomrule
    \end{tabular}
    \begin{tablenotes}
      \small
      \begin{enumerate*}
      \item[$^{1}$] With 2 $\times 23 $ subnetworks. See Table \ref{fig:namlss_structure1} for an exemplary network structure.\\ 
      \item[$^{2}$] With 23 subnetworks and each subnetwork returning a parameter for the location and shape respectively. See Table \ref{fig:namlss_structure2} for an exemplary network structure.\\ 
      \end{enumerate*}
      
    \end{tablenotes}
    \end{threeparttable}
\end{table}

\paragraph{Munich and Melbourne}

We fit the AirBnB datasets, with an Inverse Gamma distribution where applicable.  However, we train the models that only predict the mean with the squared error loss function.  While one might suspect worse performances due to that, we find that using the squared error actually leads to much smaller gamma deviances compared to the models leveraging the Inverse Gamma distribution.  Additionally, we use slightly smaller model structures than for the California Housing dataset. For all neural models, we use hidden layers of sizes $[512, 256, 50]$.  The first hidden layer is followed by a 0.5 dropout layer.  Throughout the hidden layers, we use ReLU activation functions.  However, we deviate from that for the output layer activation functions.  For the MLP we use a Softplus activation function for the output layer, ensuring that strictly positive values are predicted. For NAMLSS as well as the DNN we have to adjust the activation functions, as both models minimize the log-likelihood via the parameters $\alpha$ and $\beta$.  However, the mean prediction resulting from these parameters is defined via: 
\begin{equation*}
    \mu = \frac{\beta}{\alpha -1}
\end{equation*} and is hence only defined for $\alpha > 1$.  The activation functions thus need to ensure a $\alpha$ prediction that is larger than 1 and a $\beta$ prediction that is larger than 0.  Hence we again use a Softplus activation for the $\beta$ output layer.  For the $\alpha$ prediction, we use the following activation function element-wise:
\begin{equation*}
    h(x) = \begin{cases} \log(1+\exp(x)), & \text{if } \log(1+\exp(x)) > 1 \\
    \frac{1}{\log(1+\exp(x))}, & \text{else}.
    \end{cases}
\end{equation*}
To compute the log-likelihood for the models resulting in a mean prediction we compute the parameters $\alpha$ and $\beta$ as follows:

$$\alpha = \frac{\mu^2}{\sigma^2 + 2},$$
$$\beta = \mu \frac{\mu^2}{\sigma^2+1},$$  with $\sigma^2$ denoting the variance of the mean predictions.

For XGBoost and EBM we use a simple transformation of the target variable in order to ensure that $\mu > 0$. 
Hence we fit the model on $\log(y)$ and re-transform the predictions accordingly with $\exp(\hat{y})$. 
Otherwise, we use the same hyperparameters as for the California Housing dataset.

Interestingly, the NAM did not converge using the Softplus activation function as the MLP did. 
Using the Softplus activation resulted in tremendously large mean gamma deviances and log-likelihoods, as the model kept predicting values that were nearly zero. 
Hence, we were only able to achieve good results for the NAM using the activation function given by formula (\ref{eq:gamma_activation}).

The presented GAMLSS and gamboostLSS models assume an Inverse Gamma distribution with both $\mu$ and $\sigma$ utilizing the log-link function. 
It should be noted that the RS algorithm does not converge with GAMLSS, which is why CG is used.

\begin{table}[H]
\small
\centering
       \caption{Hyperparameters for the neural models for the Munich and Melbourne datasets}
\begin{threeparttable}
    \begin{tabular}{l|l l l l l }
        Hyperparameter      & NAMLSS$^1$ & NAMLSS$^2$ &  DNN & MLP & NAM \\
        \hline\hline
        Learning Rate       & 1e-04                 & 1e-04                 & 1e-04                 & 1e-04            & 1e-04                \\
        Dropout             & 0.5                   & 0.5                   & 0.5                   & 0.5               & 0.5                 \\
        Hidden Layers       & $[512, 256, 50]$       & $[512, 256, 50]$       & $[512, 256, 50]$       & $[512, 256, 50]$  & $[512, 256, 50]$     \\
        LR Decay, Patience  & 0.95 - 10             & 0.95 - 10             & 0.95 - 10             & 0.95 - 10        & 0.95 - 10            \\
        Activation          & ReLU                  & ReLU                  & ReLU                  & ReLU             & ReLU                 \\
        Output Activation   & Gamma$^*$, \text{Softplus}      & Gamma$^*$, \text{Softplus}      & Gamma$^*$, \text{Softplus}   & Linear         & Linear              \\
        \bottomrule
    \end{tabular}
    \begin{tablenotes}
      \small
      \begin{enumerate*}
      \item[$^{1}$] With 2 $\times 23 $ subnetworks. See Table \ref{fig:namlss_structure1} for an exemplary network structure.\\ 
      \item[$^{2}$] With 23 subnetworks and each subnetwork returning a parameter for the location and shape respectively. See Table \ref{fig:namlss_structure2} for an exemplary network structure.\\ 
      \item[$^*$] See formula (\ref{eq:gamma_activation}) for the detailed element-wise activation function. \\
      \end{enumerate*}

    \end{tablenotes}
    \end{threeparttable}
\end{table}

\end{document}